\PassOptionsToPackage{unicode}{hyperref}
\PassOptionsToPackage{hyphens}{url}
\documentclass[
]{article}
\usepackage{amsmath,amssymb}
\usepackage{lmodern}
\usepackage{iftex}
\ifPDFTeX
  \usepackage[T1]{fontenc}
  \usepackage[utf8]{inputenc}
  \usepackage{textcomp} 
\else 
  \usepackage{unicode-math}
  \defaultfontfeatures{Scale=MatchLowercase}
  \defaultfontfeatures[\rmfamily]{Ligatures=TeX,Scale=1}
\fi
\IfFileExists{upquote.sty}{\usepackage{upquote}}{}
\IfFileExists{microtype.sty}{
  \usepackage[]{microtype}
  \UseMicrotypeSet[protrusion]{basicmath} 
}{}
\makeatletter
\@ifundefined{KOMAClassName}{
  \IfFileExists{parskip.sty}{%
    \usepackage{parskip}
  }{
    \setlength{\parindent}{0pt}
    \setlength{\parskip}{6pt plus 2pt minus 1pt}}
}{
  \KOMAoptions{parskip=half}}
\makeatother
\usepackage{xcolor}
\IfFileExists{xurl.sty}{\usepackage{xurl}}{} 
\IfFileExists{bookmark.sty}{\usepackage{bookmark}}{\usepackage{hyperref}}
\hypersetup{
  pdftitle={Idiomify - Building a Collocation-supplemented Reverse Dictionary of English Idioms with Word2Vec for non-native learners},
  hidelinks,
  pdfcreator={LaTeX via pandoc}}
\usepackage[margin=1in]{geometry}
\usepackage{color}
\usepackage{fancyvrb}

\DefineVerbatimEnvironment{Highlighting}{Verbatim}{commandchars=\\\{\}}
\usepackage{framed}
\definecolor{shadecolor}{RGB}{248,248,248}
\newenvironment{Shaded}{\begin{snugshade}}{\end{snugshade}}

\newcommand{\AttributeTok}[1]{\textcolor[rgb]{0.77,0.63,0.00}{#1}}

\newcommand{\BuiltInTok}[1]{#1}

\newcommand{\CommentTok}[1]{\textcolor[rgb]{0.56,0.35,0.01}{\textit{#1}}}

\newcommand{\ControlFlowTok}[1]{\textcolor[rgb]{0.13,0.29,0.53}{\textbf{#1}}}

\newcommand{\DecValTok}[1]{\textcolor[rgb]{0.00,0.00,0.81}{#1}}

\newcommand{\KeywordTok}[1]{\textcolor[rgb]{0.13,0.29,0.53}{\textbf{#1}}}
\newcommand{\NormalTok}[1]{#1}
\newcommand{\OperatorTok}[1]{\textcolor[rgb]{0.81,0.36,0.00}{\textbf{#1}}}

\newcommand{\VariableTok}[1]{\textcolor[rgb]{0.00,0.00,0.00}{#1}}

\usepackage{longtable,booktabs,array}
\usepackage{calc} 
\usepackage{etoolbox}
\makeatletter
\patchcmd\longtable{\par}{\if@noskipsec\mbox{}\fi\par}{}{}
\makeatother
\IfFileExists{footnotehyper.sty}{\usepackage{footnotehyper}}{\usepackage{footnote}}
\makesavenoteenv{longtable}
\usepackage{graphicx}
\makeatletter
\def\maxwidth{\ifdim\Gin@nat@width>\linewidth\linewidth\else\Gin@nat@width\fi}
\def\maxheight{\ifdim\Gin@nat@height>\textheight\textheight\else\Gin@nat@height\fi}
\makeatother
\setkeys{Gin}{width=\maxwidth,height=\maxheight,keepaspectratio}
\makeatletter
\def\fps@figure{htbp}
\makeatother
\providecommand{\tightlist}{%
  \setlength{\itemsep}{0pt}\setlength{\parskip}{0pt}}
\ifLuaTeX
  \usepackage{selnolig}  
\fi

\title{Idiomify - Building a Collocation-supplemented Reverse Dictionary
of English Idioms with Word2Vec for non-native learners}
\author{Eu-Bin KIM\\
Supervised by Professor Goran Nenadic\\
Bsc Artificial Intelligence\\
Department of Computer Science\\
University of Manchester\\
\includegraphics[width=2.08333in,height=\textheight]{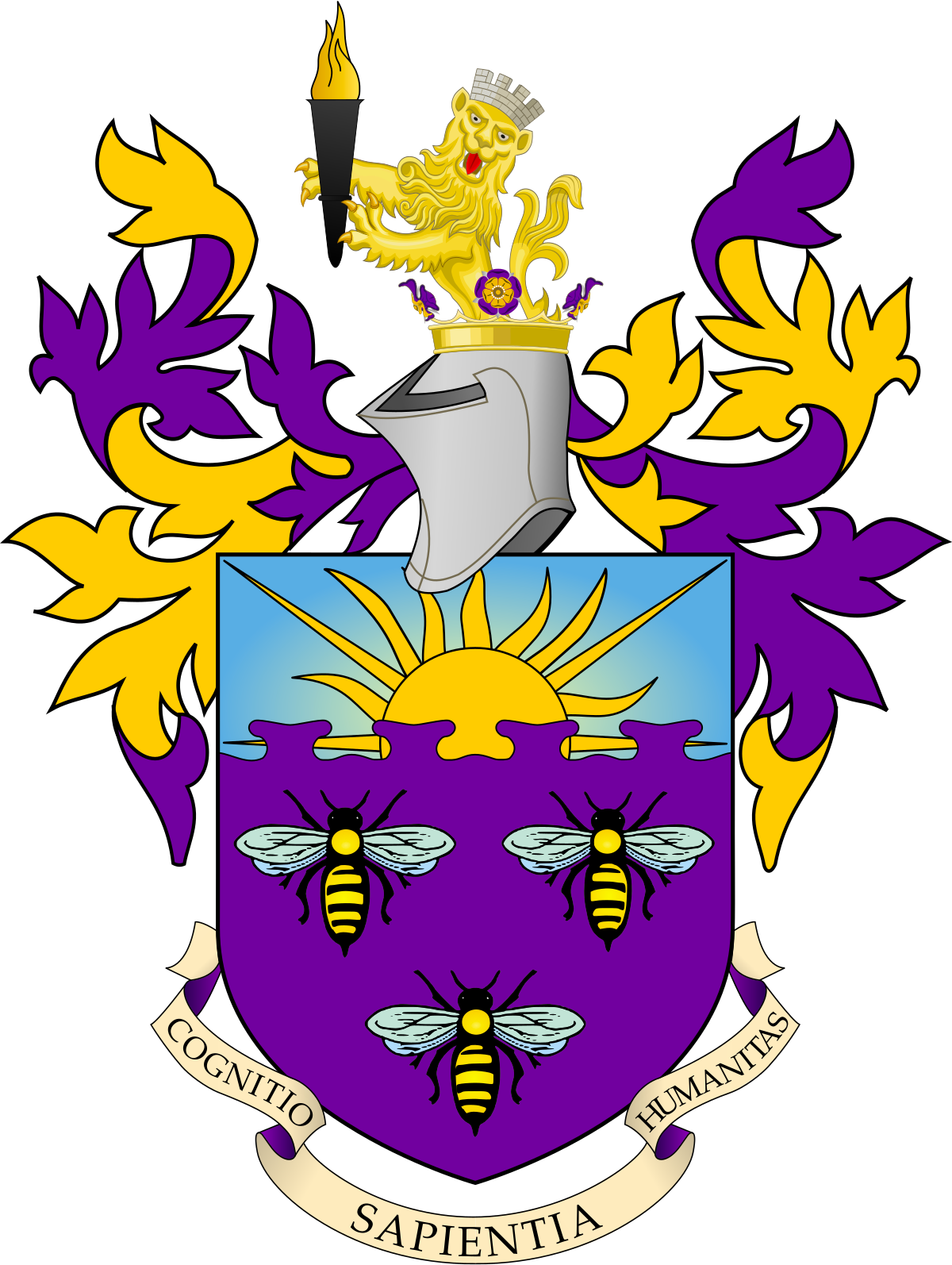}}
\date{May 2021}

\begin{document}
\maketitle

\newpage{}

\pagebreak

\hypertarget{introduction}{%
\section{1. Introduction}\label{introduction}}

\hypertarget{why-build-a-reverse-dictionary-of-idioms}{%
\subsection{1.1. Why build a reverse dictionary of
idioms?}\label{why-build-a-reverse-dictionary-of-idioms}}

\begin{longtable}[]{@{}
  >{\raggedright\arraybackslash}p{(\columnwidth - 2\tabcolsep) * \real{0.5000}}
  >{\raggedright\arraybackslash}p{(\columnwidth - 2\tabcolsep) * \real{0.5000}}@{}}
\caption{Definitions of some idioms. (Adapted from: \emph{Oxford
Advanced Learner's Dictionary of Current English}, 1995)}\tabularnewline
\toprule
\begin{minipage}[b]{\linewidth}\raggedright
idiom
\end{minipage} & \begin{minipage}[b]{\linewidth}\raggedright
definition
\end{minipage} \\
\midrule
\endfirsthead
\toprule
\begin{minipage}[b]{\linewidth}\raggedright
idiom
\end{minipage} & \begin{minipage}[b]{\linewidth}\raggedright
definition
\end{minipage} \\
\midrule
\endhead
\emph{beat around the bush} & \emph{to talk about something for a long
time without coming to the main point} \\
\emph{leave no stones unturned} & \emph{to try every possible course of
action in order to find or achieve something} \\
\emph{ring a bell} & \emph{to sound familiar to you, as though you have
heard it before} \\
\bottomrule
\end{longtable}

Idioms are figures of speech. That is, they are phrases that are more
often meant to be taken figuratively than to be taken literally (Caro \&
Edith, 2019). \textbf{Table 1} above introduces some examples of English
idioms. For instance, when people say \emph{do not beat around the
bush!}, that does not necessarily mean they are asking someone to stop
poking around the bush, but it means more to ask to get ``to the main
point''. Likewise, when people say \emph{that does not ring a bell},
this means that something does not ``sound familiar'' to them, rather
than to mean that they have a malfunctioning bell on their hands.

The figurative nature of idioms helps us communicate more effectively.
Idioms can help people describe a potentially complicated concept with a
relatively comprehensible analogy, thereby boosting their conversational
skills. For instance, It is much more comprehensible and elegant to say
\emph{They have \textbf{left stones unturned} to achieve this end} than
to rather wordly say \emph{They have \textbf{tried every possible course
of action in order} to achieve this end}. They are not just a fun way of
expressing something, they efficiently enrich the way we communicate.

Despite the benefits, non-native learners of language find it hard to
leverage idioms. This is because they significantly lack idiomatic
fluency, compared with the natives (Thyrab, 2016). As for the
non-natives, not many idioms come to their mind to begin with.
Therefore, rarely do they voluntarily try to use idioms to communicate
more effectively.

\begin{figure}
\centering
\includegraphics[width=4.16667in,height=\textheight]{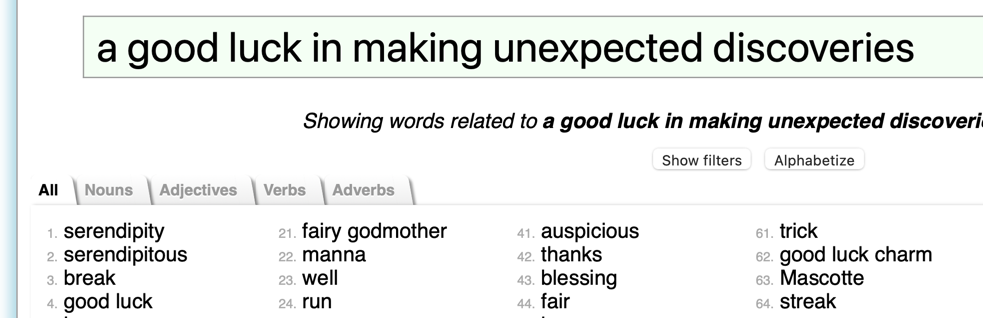}
\caption{Reverse-searching the word \emph{serendipity} with its
definition, \emph{a good luck in making unexpected discoveries} on
\emph{OneLook} (Beeferman, 2003).}
\end{figure}

We build a reverse dictionary of English idioms to help the non-native
learners of English leverage the idioms on demand. As opposed to a
forward dictionary, a reverse-dictionary allows us to search for words,
given a definition (Sierra, 2000). For example, \emph{OneLook} is a
reverse dictionary that we can use to, say, reverse-search
\emph{serendipity} with its definition, \emph{a good luck in making
unexpected discoveries} (\textbf{Figure 1}). if there was a reverse
dictionary as such but for idioms, it would be much helpful for the
non-natives, as they would be now able to explore how they could
paraphrase a sentence with an idiom, on demand. However, no
reverse-dictionaries have ever been built for idioms.

\hypertarget{why-supplement-it-with-the-collocations-of-idioms}{%
\subsection{1.2. Why supplement it with the collocations of
idioms?}\label{why-supplement-it-with-the-collocations-of-idioms}}

\begin{figure}
\centering
\includegraphics[width=4.6875in,height=\textheight]{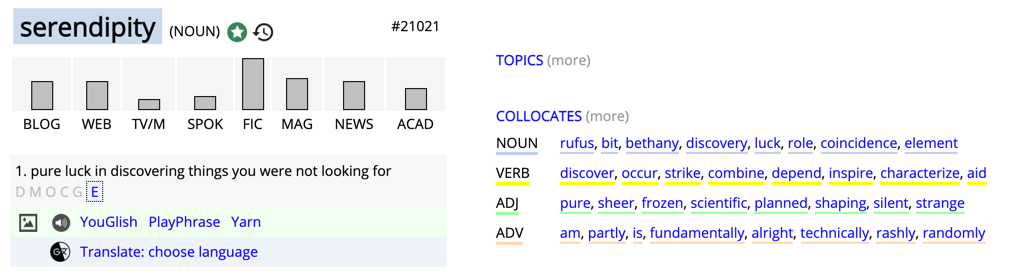}
\caption{The verb, noun, adjective and adverb collocations of the word
\emph{serendipity} (bottom right), compiled by \emph{COCA} (Davies,
2008)}
\end{figure}

Collocations are words that frequently and uniquely co-occur with each
other. Differently put, they are combination of words ``that happens
very often and more frequently than would happen by chance''
(\emph{Oxford Advanced Learner's Dictionary of Current English}, 1995)
For example, \emph{discover}, \emph{occur} and \emph{strike} are
collocates of the word \emph{serendipity} (\textbf{Figure 2} above),
because they would co-occur frequently more often \emph{serendipity}
than would co-occur by chance. Likewise, we could also have noun,
adjective and adverb collocations of words, as in \emph{rufus
serendipity}, \emph{pure serendipity} and \emph{partly serendipitous}.

\begin{longtable}[]{@{}
  >{\raggedright\arraybackslash}p{(\columnwidth - 2\tabcolsep) * \real{0.5000}}
  >{\raggedright\arraybackslash}p{(\columnwidth - 2\tabcolsep) * \real{0.5000}}@{}}
\caption{Some responses of the native and non-native speakers to
word-combining test (adapted from: Granger, 1998).}\tabularnewline
\toprule
\begin{minipage}[b]{\linewidth}\raggedright
native
\end{minipage} & \begin{minipage}[b]{\linewidth}\raggedright
non-native
\end{minipage} \\
\midrule
\endfirsthead
\toprule
\begin{minipage}[b]{\linewidth}\raggedright
native
\end{minipage} & \begin{minipage}[b]{\linewidth}\raggedright
non-native
\end{minipage} \\
\midrule
\endhead
\emph{bitterly cold}(40) & \emph{bitterly cold}(7),\emph{bitterly
aware}(3),\emph{bitterly miserable}(2) \\
\emph{blissfully happy}(19) & \emph{blissfully
happy}(4),\emph{blissfully ignorant}(20) \\
\bottomrule
\end{longtable}

Collocational knowledge is highly beneficial to non-native learners
because they serve to guide them on using words naturally and precisely.
This is partly because L2 learners tend to lack native-like heuristics
on collocation. Granger (1998) demonstrated this by having native and
non-native speakers of English take ``word-combing test'', where they
were asked to choose adjectives that acceptably collocate with a given
amplifier (e.g.~\emph{bitterly}). As \textbf{Table 2} illustrates, L2
learners showed misguided sense of collocation (e.g.~\emph{blissfully
ignorant}). This indicates that L2 learners struggle to acquire
collocational knowledge, which is why collocations dictionaries are
effective supplements to them. The editors of \emph{Oxford Collocations
Dictionary for Students of English} (2002) well exemplify the benefits;
\emph{strong rain} does get the idea across, but it would be more
natural if it were revised to \emph{heavy rain}. Likewise, \emph{a
fascinating book} is more precise than \emph{a good book} because
\emph{fascinating} collocates with \emph{book} and communicates more
than \emph{good}.

Whilst collocations of singular words are widely available, collocations
of idioms are unavailable even though they do collocate. Idioms
themselves are extremely strong collocations that have nearly fixed
forms. Therefore, it is plausible to identify idioms as a ``word'' in a
sentence, in which case collocation inevitably occurs. For example,
\emph{the government intends to seize power by hook or by crook} sounds
more natural than \emph{I'll assist you by hook or by crook}, and in
both cases \emph{by hook or by crook} effectively behaves as an atomic
unit. Yet, no major dictionary publishers have attempted to compile
collocations of idioms.

We supplement the reverse dictionary of idioms to assist the non-natives
in using idioms naturally and precisely. If we had ``Collocations
Dictionary of English Idioms'', the non-natives would benefit from it in
using idioms in an appropriate context, in a more natural way. There are
however no dictionaries as such available. This motivates us to
supplement the reverse-search engine with the collocations of retrieved
idioms.

\hypertarget{aim}{%
\subsection{1.3. Aim}\label{aim}}

\begin{figure}
\centering
\includegraphics[width=4.16667in,height=\textheight]{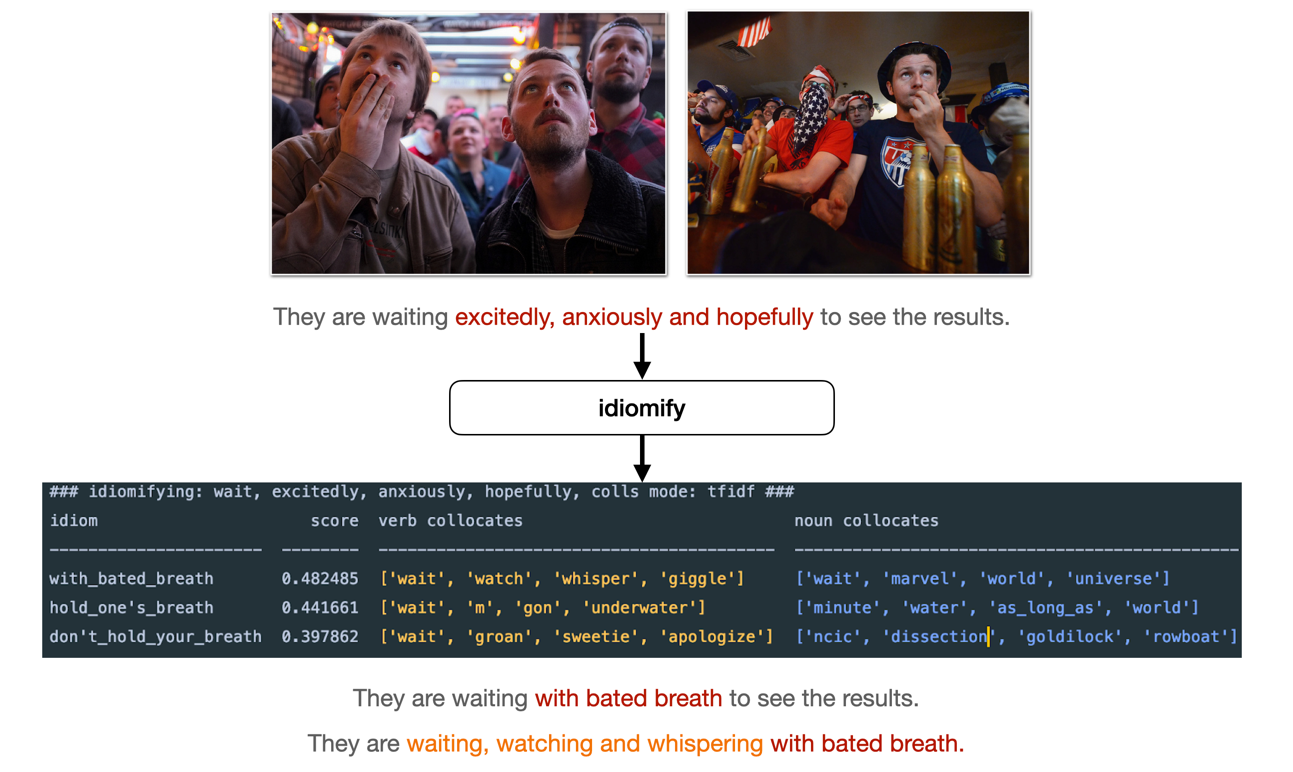}
\caption{Idiomify suggesting idioms that best describe ``wait,
excitedly, anxiously, hopefully'' altogether.}
\end{figure}

Project Idiomify aims to suggest a list of idioms that best describe a
given phrase to non-native learners of English, while supplementing the
results with the collocations of the idioms. \textbf{Figure 3}
illustrates how we might use Idiomify with an example scenario. Say we
write \emph{They are waiting excitedly, anxiously and hopefully to see
the results} to describe the people in the images above. If we were
non-native learners of English (e.g.~a Korean learning English), we may
want to explore how we could paraphrase the sentence with an idiom
(discussed in section \textbf{1.1}), as it would make the sentence more
native-like. We therefore give the phrase as the input to Idiomify:
\emph{excitedly, anxiously and hopefully}. Given the input, Idiomify
suggests \emph{with bated breath}, \emph{hold one's breath} and
\emph{don't hold your breath} as the idioms that are most likely to
capture the meaning of the phrase, of which \emph{with bated breath} is
found to be the most appropriate one. We thereby learn to rephrase the
sentence to \emph{They are waiting with bated breath to see the
results}. As being non-native learners, we may also wonder how we could
use \emph{with bated breath} more adequately than the first try
(discussed in Section \textbf{1.2.}). Here, consulting ``verb
collocates'' helps us in doing so. That is, we see that \emph{with bated
breath} collocates with \emph{watch} and \emph{whisper}, and we indeed
notice that the people in the pictures are watching something while
possibly whispering their wishful thinking. We therefore learn to revise
the first try into a more precise and communicative one: \emph{They are
waiting, watching and whispering with bated breath}.

As a result of the project, we have developed the following
deliverables:

\begin{itemize}
\tightlist
\item
  \texttt{identify-idioms}: A Python library for reliably identifying
  and collecting idiomatic expressions
\item
  \texttt{idiom2vec}: A Python library for training a Word2Vec model on
  idiomatic expressions
\item
  \texttt{idiom2colloations}: A Python library for modeling and
  extracting collocations of idioms.
\item
  \texttt{idiomify}: A python library for reverse-searching idioms
\end{itemize}

\pagebreak

\hypertarget{related-work}{%
\section{2. Related Work}\label{related-work}}

\hypertarget{identifying-idioms}{%
\subsection{2.1. Identifying Idioms}\label{identifying-idioms}}

\begin{longtable}[]{@{}
  >{\raggedright\arraybackslash}p{(\columnwidth - 4\tabcolsep) * \real{0.3333}}
  >{\raggedright\arraybackslash}p{(\columnwidth - 4\tabcolsep) * \real{0.3333}}
  >{\raggedright\arraybackslash}p{(\columnwidth - 4\tabcolsep) * \real{0.3333}}@{}}
\caption{Examples of the six variation cases of idioms. The idioms are
chosen from \emph{SLIDE} (Jochim et al., 2018).}\tabularnewline
\toprule
\begin{minipage}[b]{\linewidth}\raggedright
base form
\end{minipage} & \begin{minipage}[b]{\linewidth}\raggedright
variation
\end{minipage} & \begin{minipage}[b]{\linewidth}\raggedright
case
\end{minipage} \\
\midrule
\endfirsthead
\toprule
\begin{minipage}[b]{\linewidth}\raggedright
base form
\end{minipage} & \begin{minipage}[b]{\linewidth}\raggedright
variation
\end{minipage} & \begin{minipage}[b]{\linewidth}\raggedright
case
\end{minipage} \\
\midrule
\endhead
\emph{balls-out} & \emph{balls out} & optional hyphen \\
\emph{find one's feet} & \emph{finding her feet} & inflection \\
\emph{add fuel to the fire} & \emph{throw gasoline on the fire} &
alternatives \\
\emph{grasp at straws} & \emph{grasp desperately at straws} &
modification \\
\emph{open the floodgates} & \emph{the floodgates were opened} &
passivisation \\
\emph{keep someone at arm's length} & \emph{keep his old friends at
arm's length} & open slot \\
\bottomrule
\end{longtable}

It is challenging to identify idioms because they extensively vary in
forms. We need a reliable way of identifying idioms to collect as many
idiomatic expressions from corpora. However, while some idioms are
syntactically frozen (e.g.\emph{by hook or by crook}), many of the
idioms vary. We could categorize their variations into 6 cases: optional
hyphen, inflection, alternatives, modification, open slot and
passivisation. \textbf{Table 3} exemplifies each case. We often omit the
hyphen (e.g.~\emph{balls out}), the verbs and possessive pronouns
inflect (e.g.~\emph{finding her feet}), some idioms have alternative
forms (e.g.~\emph{throw gasoline on the fire} ). In addition to these
cases, Hughs et al.~point out (2021) that there are three more cases;
some constituent words could be modified with modals ( e.g.~\emph{grasp
desperately at straws}), idioms could be passivised (e.g.~\emph{the
floodgates were opened}), pronouns could be substituted with a set of
words (e.g.~\emph{keep his old friends at arm's length}). We can by no
means reliably identify idioms unless we come up with a way of
addressing all of these variations.

At the late stages of this project, Hughs et al.~have published (2021)
their approach on identifying idioms. They do so by leveraging a search
query API provided by a flexible and scalable search engine called
ElasticSearch (Gormley \& Tony, 2015); As will be discussed in later
sections, they generate a set of queries for each idiom, each of which
is specifically designed to address the aforementioned six variation
cases, and hence their paper's name: ``leaving no stones unturned''.

The set of idioms offered by \emph{SLIDE} project is worthy of
extracting the collocations for and building a reverse-dictionary for.
This is because \emph{SLIDE} offers a shortlist of frequently used
idioms (Jochim et al., 2018). They first collected 8772 idioms from
Wiktionary, then compiled only the 5000 most frequently occurring idioms
in news articles, publications, etc. Therefore, one could get the most
out of their exploration on idioms if they do so with the set of idioms
included in \emph{SLIDE}.

\hypertarget{modeling-collocations-of-idioms}{%
\subsection{2.2. Modeling collocations of
Idioms}\label{modeling-collocations-of-idioms}}

Researchers have attempted to model the notion of collocations with
statistical metrics that can measure pairwise significance. That is, if
a metric can statistically determine that, for example, the pairwise
significance of \emph{heavy rain} is larger than that of \emph{strong
rain}, people have used the metric to mathematically model collocations.
This is a reasonable approach, as the notion of pairwise significance as
such well echoes with the aforementioned definition of collocations -
``Words that uniquely and frequently co-occur each other'' (Section
\textbf{1.2}). The pairwise significance metrics that have been used to
model collocations include: T-score, Pearson's X-square Test,
Log-likelihood Ratio and Pointwise Mutual Information (Thanopoulos et
al., 2002).

\begin{quote}
\[PMI(w1, w2) = log_{2}\frac{P(w_{1}, w_{2})}{P(w_{1}) * P(w_{2})}\]
\textbf{Equation 1} The equation of Pointwise Mutual Information (PMI)
for bigrams. (adapated from: Church \& Hanks, 1989)
\end{quote}

Point-wise Mutual Inclusion (PMI) is one of the most widely used measure
for modeling collocation. First proposed by Church \& Hanks in 1989, PMI
has long been used to model collocations and extract them from corpora.
\textbf{Equation 1} shows the mathematical definition of PMI. Here, we
see that PMI is proportional to the \(P(w_{1}, w_{2})\) conjugation
probability. This means that the more frequently the words co-occur in a
corpus, the higher the PMI score gets. We also see that PMI is inversely
proportional to the product of \(P(w_{1})\) and \(P(w_{2})\). This means
that the more rarely each word independently occurs in a corpus, the
higher the PMI score gets. All in all, PMI favours rarely but frequently
co-occurring pairs of words, which is a good fit for collocations.

\begin{quote}
\[tf*idf_{t, d} = (1 + log_{10}tf_{t,d}) * log_{10}(N/df_{t})\]
\textbf{Equation 2} : The equation of TfIdf weighting (adapted from:
Manning et al., 2008)
\end{quote}

Term Frequency - Inverse Document Frequency (TF-IDF) is also a popular
metric for pairwise significance, which could be used to model
collocations. TF-IDF has long been used in information retrieval to
measure the significance (the weight) of a word in a specific document,
so that the weight can be used to rank documents (Manning et al., 2008).
In the regime of TF-IDF weighting (\textbf{Equation 2}), the weight of a
term \texttt{t} is proportional to its term-frequency \texttt{tf} in a
document \texttt{d} and inversely proportional to its document-frequency
\texttt{df} in \texttt{d}. In qualitative terms, this means that the
more \texttt{t} frequently and uniquely co-occurs with \texttt{d}, the
higher tf-idf weight of \texttt{t} with respect to \texttt{d} gets.
Here, if we think of \texttt{t} as a word and \texttt{d} as another word
that co-occurs with the word \texttt{t}, we can see that TF-IDF is
conceptually just as suitable model for collocations as PMI. Extracting
collocations from corpora is retrieving documents and ranking them in
terms of their collocative significance after all, so it is worthy of
exploring TF-IDF as a candidate for the model of collocations.

\hypertarget{reverse-searching-idioms}{%
\subsection{2.3. Reverse-searching
Idioms}\label{reverse-searching-idioms}}

\begin{longtable}[]{@{}
  >{\raggedright\arraybackslash}p{(\columnwidth - 2\tabcolsep) * \real{0.4712}}
  >{\raggedright\arraybackslash}p{(\columnwidth - 2\tabcolsep) * \real{0.5288}}@{}}
\caption{An illustration of the difference between a forward index and
an inverted index. (Adapted from: Ingebrigsten, 2013)}\tabularnewline
\toprule
\begin{minipage}[b]{\linewidth}\raggedright
forward index
\end{minipage} & \begin{minipage}[b]{\linewidth}\raggedright
inverted index
\end{minipage} \\
\midrule
\endfirsthead
\toprule
\begin{minipage}[b]{\linewidth}\raggedright
forward index
\end{minipage} & \begin{minipage}[b]{\linewidth}\raggedright
inverted index
\end{minipage} \\
\midrule
\endhead
\emph{Grandma's tomato soup}: leaves, tomatoes, basil & basil:
\emph{Grandma's tomato soup} \\
\emph{African tomato soup}: tomatoes, leaves, baobab & leaves:
\emph{African tomato soup}, \emph{Grandma's tomato soup} \\
\emph{Good ol' tomato soup}: tomato, salt, garlic & salt: \emph{African
tomato soup}, \emph{Good ol' tomato soup} \\
\bottomrule
\end{longtable}

People have taken three approaches to engineering a reverse-dictionary
of words: inverted index, graphs and distributional semantics. The
simplest approach of all is populating a def-to-word inverted index. As
opposed to a forward index, an inverted index maps a term to documents,
which allows us to search documents by terms (Ingebrigsten, 2013). For
instance, as illustrated in \textbf{Table 4} above, the inverted index
of recipes allows us to search them by their ingredients
(e.g.~\emph{leaves} -\textgreater{} \emph{African tomato soup}). If we
think of words as the recipes and each term in their definitions as the
ingredients, if we populate a def-to-word inverted index that is, we can
conceivably implement a simple reverse-dictionary of words.
\emph{OneLook}, introduced in Section \textbf{1.1}, is an example of a
reverse-dictionary that takes this approach. While \emph{OneLook} works
fairly well, Thorat \& Choudhari. attempted (2016) to improve upon
\emph{OneLook} by taking a graph-based approach. Their idea was to
leverage \emph{WordNet} (Miller, 1995), from which they extracted
hypernym-hyponym relations of words (e.g.~\emph{parent} is a hypernym
for \emph{father}), to search the word that is nearest to all the terms
of a given phrase. For example, as illustrated in \textbf{Figure 4}
below, on searching their graph with the phrase \emph{Son of my
parents}, the word \emph{brother} would be retrieved as it is the one
that is closest to both \emph{Son} and \emph{Parents} with respect to
their similarity measure. It has also been attempted to take a
distributional semantics approach. Distributional semantics assumes that
the semantics are distributed across contexts. That is, they aim to find
the meaning of words from a corpus rather than a dictionary. For
example, Hill et al.~attempts (2016) to build a reverse dictionary by
having a Recurrent Neural Network learn the mapping from definition to
words. Unlike how Thorat \& Choudhari et al.~carefully engineered a way
to leverage \emph{Word2Net}, Hill et al.~simply provide their model
features and labels. And yet, their RNN turns out to perform just as
good as \emph{OneLook}.

\begin{figure}
\centering
\includegraphics[width=3.125in,height=\textheight]{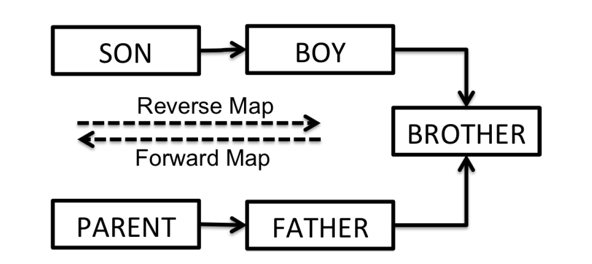}
\caption{The reverse map leading the phrase \emph{Son of my parents} to
the word \emph{brother}. The hypernym-hyponym relations of the words are
extracted from \emph{WordNet}. (Adapted from: Thorat \& Choudhari,
2016)}
\end{figure}

The distributional semantics approach is the most flexible approach of
all. One reason for this is that they do not require a domain-specific
feature engineering. This is especially useful for building a
reverse-dictionary of idioms, since we don't have any ``IdiomNet'' to
leverage features from. Another reason for the flexibility is that
distributional semantics allows for a semantics-aware search. For
instance, Hill et al.'s aforementioned RNN model can reverse-search
words not only given a definition, but also given a concept description;
their model can find the word \emph{prefer} given \emph{when you like
one thing more than another thing}, although this is a paraphrased
version of the definition of \emph{prefer}.

Word2Vec is a good baseline for the distributional approach to building
a reverse-dictionary. In their landmark paper, Mikolov et al.~(2013)
devised a text-based language model that can work out word analogies.
The model, now known as Word2Vec, was capable of finding X such that
\emph{Woman is to Man as Queen is to X}, in which case it would output a
list words related to \emph{King}. Perhaps due to this remarkable of
numerically capturing semantics of words, Hill et al.~also uses (2016)
Word2Vec to set up a baseline model for their reverse-dictionary. They
use Word2Vec to get the word embeddings vectorize a phrase by averaging
the word embeddings of each word in the phrase. Despite its
simplicity,this way of Word2Vec-averaging a sentence is a ``A simple but
tough-to-beat baseline for sentence embeddings'' (Arora et al., 2017).
Hence, it would not be a terrible idea to implement a reverse-dictionary
by first Word2Vec-average an input phrase, and then finding the nearest
word neighbours to the average vector.

\pagebreak

\hypertarget{implementation}{%
\section{3. Implementation}\label{implementation}}

\begin{figure}
\centering
\includegraphics[width=4.6875in,height=\textheight]{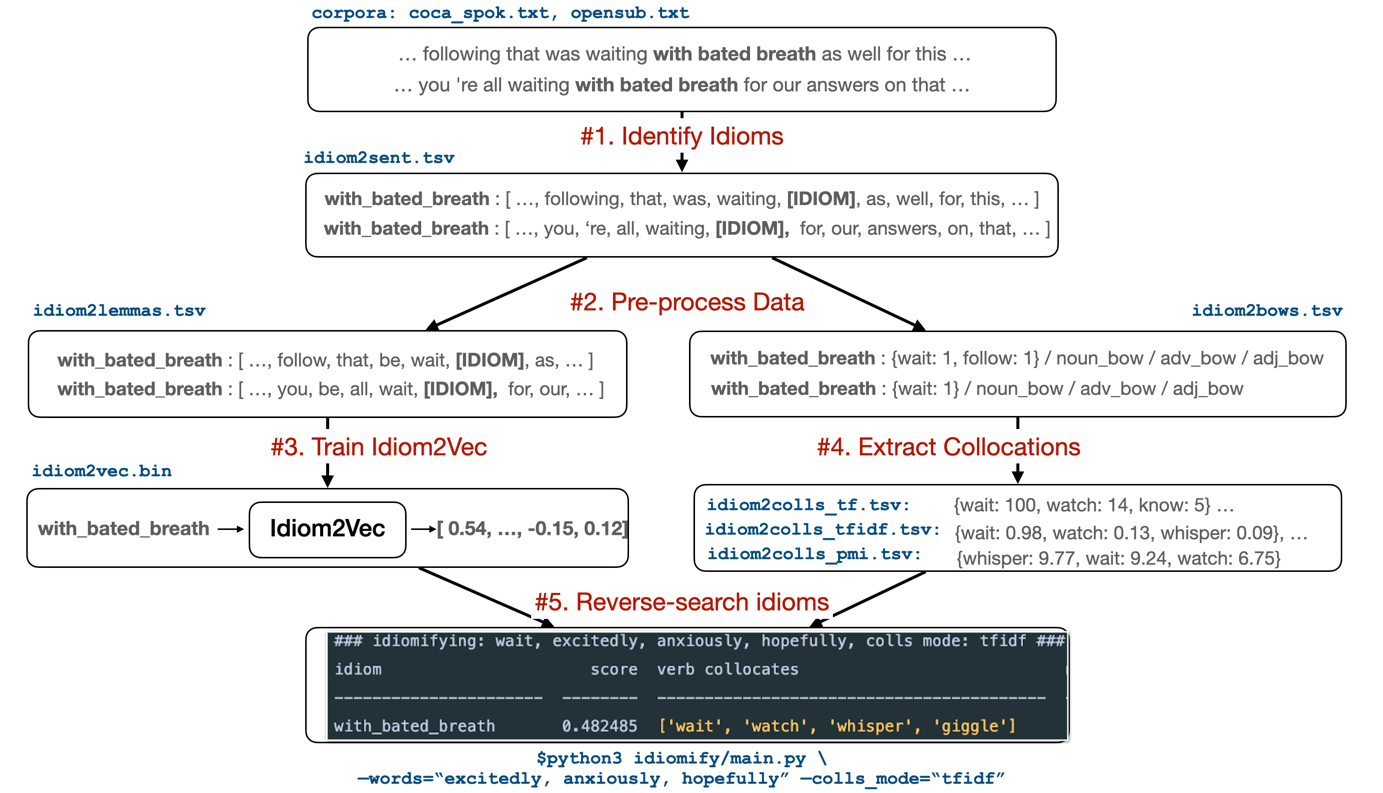}
\caption{The five subgoals of project Idiomify with concrete examples}
\end{figure}

As overviewed in \textbf{Figure 5} above, the main goal of project
Idiomify is broken down into the following five subgoals:

\begin{enumerate}
\def\labelenumi{\arabic{enumi}.}
\tightlist
\item
  \textbf{identify idioms}: identify and collect idiomatic expressions
  from corpora
\item
  \textbf{pre-process data}: prepare the data needed for step 3, 4 and
  5, from the data collected in step 1.
\item
  \textbf{extract collocations of idioms}: extract the collocations of
  idioms with TF, PMI and TFIDF from the data processed in step 2.
\item
  \textbf{train an Idiom2Vec Model}: train a Word2Vec model on the data
  processed in step 2.
\item
  \textbf{reverse-search idioms (Idiomify)}: idiomify a given phrase
  using the Idiom2Vec trained in step 4.
\end{enumerate}

In this chapter, we elaborate on how we implement each step.

\hypertarget{identifying-idioms-1}{%
\subsection{3.1. Identifying idioms}\label{identifying-idioms-1}}

We get a vocabulary of idioms to identify from \emph{SLIDE} project.
This is because it offers 5000 most frequently used idioms, as discussed
in Section \textbf{2.1}. We however do not identify all the 5000 idioms;
we exclude the short ones. This is because they either hold little
significance or often used literally rather than figuratively
(e.g.~\emph{I do}, \emph{I wish} , \emph{add up}, etc). However, we
include hyphenated idioms in the vocabulary regardless of their lengths
because they are almost always taken figuratively (e.g.~\emph{catch-22},
\emph{need-to-know}, \emph{back-to-back}, etc). All things considered,
this leaves us with 2746 idioms to identify, which is still a large set
of idioms to explore. The list of all the identifiable idioms is
viewable on \texttt{identify-idioms} library, which we will discuss at
the end of this section.

\begin{longtable}[]{@{}
  >{\raggedright\arraybackslash}p{(\columnwidth - 2\tabcolsep) * \real{0.5000}}
  >{\raggedright\arraybackslash}p{(\columnwidth - 2\tabcolsep) * \real{0.5000}}@{}}
\caption{Examples of the matching rules for identifying idioms. Three
variation cases are covered: optional hyphen, alternatives and
inflection.}\tabularnewline
\toprule
\begin{minipage}[b]{\linewidth}\raggedright
base form / variation case
\end{minipage} & \begin{minipage}[b]{\linewidth}\raggedright
matching rule
\end{minipage} \\
\midrule
\endfirsthead
\toprule
\begin{minipage}[b]{\linewidth}\raggedright
base form / variation case
\end{minipage} & \begin{minipage}[b]{\linewidth}\raggedright
matching rule
\end{minipage} \\
\midrule
\endhead
\emph{down-to-earth} / optional hyphen &
\texttt{{[}{[}{[}TEXT:down{]};\ {[}TEXT:to{]};\ {[}TEXT:earth{]}{]},\ {[}{[}TEXT:down{]};\ {[}TEXT:-{]};\ {[}TEXT:to{]};\ {[}TEXT:-{]};\ {[}TEXT:earth{]}{]}{]}} \\
\emph{add insult to injury} / alternatives &
\texttt{{[}{[}{[}LEMMA:add{]};\ {[}LEMMA:insult{]};\ {[}LEMMA:to{]};\ {[}LEMMA:injury{]}{]},\ {[}{[}LEMMA:heap{]};\ {[}LEMMA:insult{]};\ {[}LEMMA:on{]};\ {[}LEMMA:injury{]}{]}{]}} \\
\emph{find one's feet} / inflection &
\texttt{{[}{[}LEMMA:find{]};\ {[}POS:\$PRP{]};\ {[}LEMMA:feet{]}{]}} \\
\bottomrule
\end{longtable}

We handle the case of optional hyphen, inflection and alternatives by
defining idiom-matching rules that address each case. We do this by
automatically deriving matching rules from the base forms of each idiom,
an illustration of which is shown in \textbf{Table 5}. As for hyphenated
idioms, we logical-or the rules with and without the hyphen (denoted as
\texttt{,}), each of which is the constituent texts sequentially joined
with logical-and, denoted as \texttt{;} (e.g.~\emph{down-to-earth}).
Similarly, as for the idioms with alternatives, we logical-or all the
possible alternatives, each of which is the constituent lemmas
sequentially joined with logical-and. Here, defining the rules in terms
of \texttt{LEMMA} ensures the rules to cover any possible inflection of
the constituent words. As for the idioms with inflecting personal
pronouns, we include a rule that is defined in terms of part-of-speech
(e.g.~\texttt{{[}POS:\$PRP{]}} ). The automatic derivation of the
matching rules were made possible with minimal boilerplate code by
\texttt{spaCy}'s (Honnibal \& Montani, 2017) easy-to-use
\texttt{Language} pipeline and \texttt{Matcher} API.

\begin{longtable}[]{@{}ll@{}}
\caption{Example entries of \texttt{idiom2sent.tsv} with the original
sentences.}\tabularnewline
\toprule
column 1 (idiom) & column 2 (tokens) \\
\midrule
\endfirsthead
\toprule
column 1 (idiom) & column 2 (tokens) \\
\midrule
\endhead
\emph{down-to-earth} &
\texttt{{[}She,\ \textquotesingle{}s,\ so,\ cool,\ and,\ {[}IDIOM{]}{]}} \\
\emph{add insult to injury} &
\texttt{{[}For,\ God,\ \textquotesingle{}s,\ sake,\ man,\ do,\ not,\ {[}IDIOM{]}{]}} \\
\emph{find one's feet} &
\texttt{{[}drift,\ back,\ in,\ time,\ and,\ {[}IDIOM{]}{]}} \\
\bottomrule
\end{longtable}

With the matching rules, a Python library, \texttt{identify-idioms}, was
developed and used to collect idiomatic expressions from two corpora.
The library is published with \texttt{pip} for open use. Hence, anyone
can install it ( \texttt{pip3\ install\ identify-idioms}) and use it to
identify idioms as atomic tokens. We use the library to identify idioms
from two corpora - the spoken data of \emph{Corpus of Contemporary
American English}, otherwise known as \emph{COCA}(Davis, 2008), and
OpenSubtitles (Tiedmann, 2012). This is because they are both rich with
spoken contexts (e.g.~movie subtitles), wherein idiomatic expressions
are frequently uttered. As such, we were able to collect more than 3
million idiomatic utterances from the two corpora. We save this into a
tab-separated values named \texttt{idiom2sent.tsv}, some examples of
which are represented in \textbf{Table 6} above.

\hypertarget{pre-processing-data}{%
\subsection{3.2. Pre-processing data}\label{pre-processing-data}}

\begin{longtable}[]{@{}
  >{\raggedright\arraybackslash}p{(\columnwidth - 2\tabcolsep) * \real{0.5000}}
  >{\raggedright\arraybackslash}p{(\columnwidth - 2\tabcolsep) * \real{0.5000}}@{}}
\caption{Some example entries of
\texttt{idiom2lemma2pos.tsv}}\tabularnewline
\toprule
\begin{minipage}[b]{\linewidth}\raggedright
col 1 (idiom)
\end{minipage} & \begin{minipage}[b]{\linewidth}\raggedright
col 2 (lemma-pos pairs)
\end{minipage} \\
\midrule
\endfirsthead
\toprule
\begin{minipage}[b]{\linewidth}\raggedright
col 1 (idiom)
\end{minipage} & \begin{minipage}[b]{\linewidth}\raggedright
col 2 (lemma-pos pairs)
\end{minipage} \\
\midrule
\endhead
out\_of\_touch &
\texttt{{[}{[}o\textquotesingle{}reilly,\ X{]},\ {[}well,\ INTJ{]},\ {[}then,\ ADV{]},\ {[}you,\ PRON{]},\ {[}be,\ VERB{]},\ {[}{[}IDIOM{]},\ X{]}{]}} \\
on\_one's\_toes &
\texttt{{[}{[}they,\ PRON{]},\ {[}keep,\ VERB{]},\ {[}you,\ PRON{]},\ {[}{[}IDIOM{]},\ X{]}{]}} \\
get\_to\_the\_point &
\texttt{{[}{[}let,\ VERB{]},\ {[}\textquotesingle{}s,\ PRON{]},\ {[}{[}IDIOM{]},\ X{]}{]}} \\
\bottomrule
\end{longtable}

In Pre-processing step, we prepare the data needed for training an
Idiom2Vec. That is, we lemmatise and pos-tag the set of idiomatic
expressions, \texttt{idiom2sent.tsv}. The result of this is saved in
\texttt{idiom2lemma2pos.tsv}, some examples of which are shown in
\textbf{Table 7} above. This is later used to produce
\texttt{idiom2bows.tsv} (discussed in next paragraph), and to train
Idiom2Vec in step 4 (discussed in Section \textbf{3.4}.

\begin{longtable}[]{@{}
  >{\raggedright\arraybackslash}p{(\columnwidth - 2\tabcolsep) * \real{0.5000}}
  >{\raggedright\arraybackslash}p{(\columnwidth - 2\tabcolsep) * \real{0.5000}}@{}}
\caption{Some example entries of \texttt{idiom2bows.tsv}}\tabularnewline
\toprule
\begin{minipage}[b]{\linewidth}\raggedright
col 1 (idiom)
\end{minipage} & \begin{minipage}[b]{\linewidth}\raggedright
col 2 (verb bows) / col 3 (noun bows) / col 4 (adj bows) / col 5 (adv
bows)
\end{minipage} \\
\midrule
\endfirsthead
\toprule
\begin{minipage}[b]{\linewidth}\raggedright
col 1 (idiom)
\end{minipage} & \begin{minipage}[b]{\linewidth}\raggedright
col 2 (verb bows) / col 3 (noun bows) / col 4 (adj bows) / col 5 (adv
bows)
\end{minipage} \\
\midrule
\endhead
lose\_one's\_mind &
\texttt{\{\}\ /\ \{\}\ /\ \{\}\ /\ \{completely:\ 3\}} \\
the\_thing\_is &
\texttt{\{know:\ 2\}\ /\ \{\}\ /\ \{important:\ 3\}\ /\ \{\}} \\
you\_know\_what &
\texttt{\{think:\ 3\}\ /\ \{sociology:\ 3\}\ /\ \{\}\ /\ \{\}} \\
left\_and\_right &
\texttt{\{comment:\ 3,\ happen:\ 2\}\ /\ \{doctor:\ 2\}\ /\ \{bizarre:\ 3\}\ /\ \{\}} \\
in\_the\_hospital &
\texttt{\{suffer:\ 2\}\ /\ \{\}\ /\ \{\}\ /\ \{financially:\ 2\}} \\
\bottomrule
\end{longtable}

We also prepare the data to extract the collocations of idioms from.
That is, we collect verb, noun, adjective and adverb lemmas that are in
vicinity to idioms from \texttt{idiom2lemma2pos.tsv}. We process them
into a bag-of-words dictionaries, the result of which is saved in
\texttt{idiom2bows.tsv} (e.g.~\textbf{Table 8}). This data is later used
as the source for collocations of idioms, from which we extract
collocations in the following step 3.

\hypertarget{modeling-and-extracting-the-collocations-of-idioms}{%
\subsection{3.3. Modeling and extracting the collocations of
idioms}\label{modeling-and-extracting-the-collocations-of-idioms}}

\begin{figure}
\centering
\includegraphics[width=4.42708in,height=\textheight]{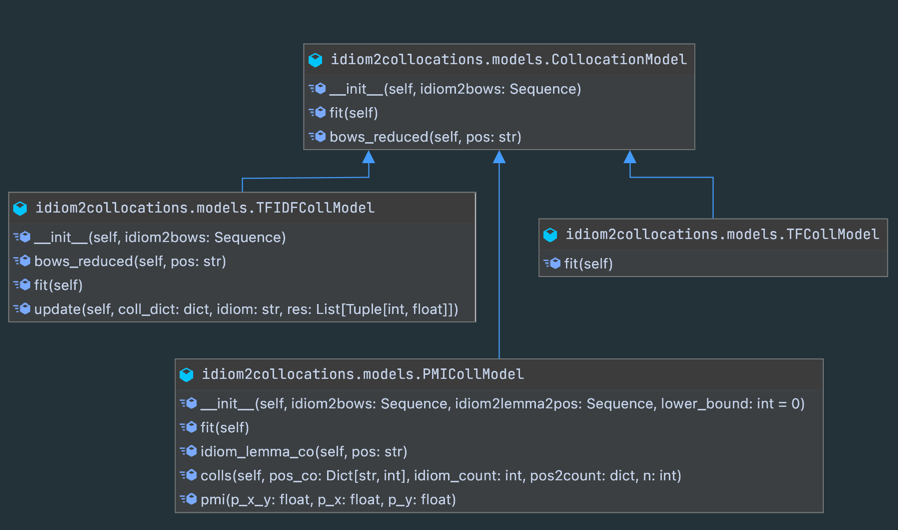}
\caption{The UML diagram of the collocation model classes that are
implemented. There are three models of collocations:
\texttt{TFCollModel}, \texttt{TFIDFCollModel} and \texttt{PMICollModel},
all of which are subclasses of \texttt{CollModel}. Given a data, they
extract the collocations of idioms on \texttt{fit()} method call.}
\end{figure}

We model the collocations with PMI, TF-IDF and Term Frequency (TF). The
reason for modeling collocations with PMI and TF-IDF is because PMI has
been widely used to model collocations, and TF-IDF is also a popular
metric of pairwise significance that we can conceivably use to model
collocations, as discussed in Section \textbf{2.2}. In addition, we also
explore Term Frequency (TF) to have a reasonable baseline with which we
can compare the other two models. The three models are implemented as
Python classes, the UML diagram of which is illustrated in
\textbf{Figure 6}. Given \texttt{idiom2bows.tsv} as the input, the three
models extract four types of collocations: noun, verb, adjective and
adverb collocations. This is to produce the same outcome as the
collocations offered by \emph{COCA}, which was introduced in Section
\textbf{1.2}.

\begin{quote}
\begin{Shaded}
\begin{Highlighting}[]
\AttributeTok{@staticmethod}
\KeywordTok{def}\NormalTok{ pmi(p\_x\_y: }\BuiltInTok{float}\NormalTok{, p\_x: }\BuiltInTok{float}\NormalTok{, p\_y: }\BuiltInTok{float}\NormalTok{) }\OperatorTok{{-}\textgreater{}} \BuiltInTok{float}\NormalTok{:}
    \CommentTok{"""}
\CommentTok{    Point{-}wise Mutual Information}
\CommentTok{    """}
\NormalTok{    log\_p\_x\_y }\OperatorTok{=}\NormalTok{ math.log(p\_x\_y, }\DecValTok{2}\NormalTok{)}
\NormalTok{    log\_p\_x }\OperatorTok{=}\NormalTok{ math.log(p\_x, }\DecValTok{2}\NormalTok{)}
\NormalTok{    log\_p\_y }\OperatorTok{=}\NormalTok{ math.log(p\_y, }\DecValTok{2}\NormalTok{)}
    \ControlFlowTok{return}\NormalTok{ log\_p\_x\_y }\OperatorTok{{-}}\NormalTok{ (log\_p\_x }\OperatorTok{+}\NormalTok{ log\_p\_y)}
\end{Highlighting}
\end{Shaded}

Code 1: The python implementation of PMI. This is one of the methods of
\texttt{PMICollModel} class.
\end{quote}

We implement \texttt{PMICollModel} by writing the definition of PMI from
scratch, and implement \texttt{TFIDFCollModel} with \texttt{Gensim}
library. \textbf{Code 1} shows the python implementation of PMI, which
is the core method of the \texttt{PMICollModel} class. Although the
original definition of PMI is \(PMI(x,y) = log(p(x,y)/(p(x)p(y)))\), we
use the product \& quotient rule of logarithms
(i.e.~\(log(a * b) = log(a) + log(b)\) and
\(log(a / b) = log(a) - log(b)\)) to translate it into additions and
subtractions: \(log(p(x, y)) - (log(p(x)) + log(P(y)))\), for faster and
more precise computation of PMI. As for our \texttt{TFIDFCollModel}
class, it is implemented with \texttt{Gensim} (Rehurek \& Sojka, 2011)
library's \texttt{TfidfModel} API.

A python library, \texttt{idiom2colloations}, was developed to house the
collocation model classes and use them to extract collocations. Given
\texttt{idiom2bows.tsv} as the input, \texttt{TFCollModel},
\texttt{TFIDFCollModel} and \texttt{PMICollModel} extract the
collocations of idioms and save them to three tsv files:
\texttt{idiom2colls\_tf.tsv}, \texttt{idiom2colls\_tfidf.tsv},
\texttt{idiom2colls\_pmi.tsv}. These files are used to compile top 5
collocations of idioms, some examples of which can be found in Appendix
\textbf{1A}, \textbf{2A}, \textbf{3A} and \textbf{4A}.

\hypertarget{train-idiom2vec}{%
\subsection{3.4. Train Idiom2Vec}\label{train-idiom2vec}}

We choose to use Word2Vec for implementing the reverse-dictionary of
idioms. This is because distributional semantics approach is flexible,
for which Word2Vec is a simple yet strong baseline, as discussed in
Section \textbf{2.3}.

\begin{longtable}[]{@{}lllll@{}}
\caption{The three versions of Idiom2Vec that were trained. each model
was trained on a slightly different training set.}\tabularnewline
\toprule
Idiom2Vec & corpora & stopwords & lemmatisation & \\
\midrule
\endfirsthead
\toprule
Idiom2Vec & corpora & stopwords & lemmatisation & \\
\midrule
\endhead
V1 & \emph{COCA} (spok) & not removed & lemmatised & \\
V2 & \emph{COCA} (spok) + \emph{OpenSubtitles} & not removed &
lemmatised & \\
V3 & \emph{COCA} (spok) + \emph{OpenSubtitles} & removed & lemmatised
& \\
\bottomrule
\end{longtable}

In order to get dense vector representations of idioms, we train
Word2Vec models on a set of idiomatic expressions. Since the goal is to
vectorize idioms, we call them ``Idiom2Vec''. We train Idiom2Vec on
different versions of \texttt{idiom2lemma2pos.tsv} (obtained from step
2) to explore the effect of corpus size and stopwords. We therefore have
three versions of Idiom2Vec, as shown in \textbf{Table 9}. V1 is trained
on \emph{COCA} only, while V2 and V3 are trained on \emph{OpenSubtitles}
as well as \emph{COCA}. This is to see if increasing the size of the
training data results in any improvement on the performance of Idiomify
(the last step). Stopwords are a set of words that hold little semantic
significance (Ganesan, 2020), which are generally filtered out to focus
on the meaningful words. However, whether they are useless or not is
domain-dependent. We therefore do not remove stopwords for training V1
and V2, but remove them for training V3, to experiment on the effect of
stopwords on the performance of Idiomify. Training \& serialising the
three models were implemented with \texttt{Gensim} (Rehurek \& Sojka,
2011) library's \texttt{Word2Vec} API.

\begin{figure}
\centering
\includegraphics[width=3.64583in,height=\textheight]{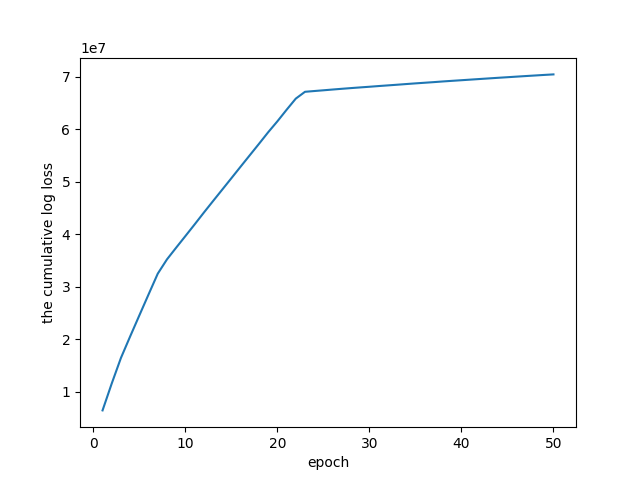}
\caption{The cumulative log loss of Idiom2Vec V1 over 50 epochs.}
\end{figure}

\begin{longtable}[]{@{}lllllll@{}}
\caption{The hyper parameters that were used to train IdiomVec V1, V2
and V3.}\tabularnewline
\toprule
model & vector & epoch & window & min count & learning rate & type \\
\midrule
\endfirsthead
\toprule
model & vector & epoch & window & min count & learning rate & type \\
\midrule
\endhead
V1 & 200 & 50 & 8 & 1 & 0.025 & Skipgram \\
V2 & 200 & 80 & 8 & 1 & 0.025 & Skipgram \\
V3 & 200 & 80 & 8 & 1 & 0.025 & Skipgram \\
\bottomrule
\end{longtable}

Aside from epoch, the three models are trained with the same set of
hyper parameters, as shown in \textbf{Table 10}. Ideally, all of them
should be optimised to each model. This however was not feasible as
training the models would take over four hours. Therefore, as for vector
size, window size, min count, learning rate and the type of Word2Vec, we
fix them with their conventional values. We however do optimise epoch by
monitoring the cumulative log loss. That is, we stop the training as
soon as the cumulative loss starts to plateau. For example, the number
of epochs for V1 is set to 50 (\textbf{Table 10}), because the loss of
V1 starts to plateau at 50 epoch, as illustrated in \textbf{Figure 7}.

\begin{longtable}[]{@{}
  >{\raggedright\arraybackslash}p{(\columnwidth - 4\tabcolsep) * \real{0.3333}}
  >{\raggedright\arraybackslash}p{(\columnwidth - 4\tabcolsep) * \real{0.3333}}
  >{\raggedright\arraybackslash}p{(\columnwidth - 4\tabcolsep) * \real{0.3333}}@{}}
\caption{The top 5 nearest idioms to \emph{catch-22}. The results are
obtained with Idiom2Vec V2. The definitions are adapted from
\emph{Oxford Learner's Dictionary of English} (1995).}\tabularnewline
\toprule
\begin{minipage}[b]{\linewidth}\raggedright
rank / idiom
\end{minipage} & \begin{minipage}[b]{\linewidth}\raggedright
cosine similarity
\end{minipage} & \begin{minipage}[b]{\linewidth}\raggedright
definition
\end{minipage} \\
\midrule
\endfirsthead
\toprule
\begin{minipage}[b]{\linewidth}\raggedright
rank / idiom
\end{minipage} & \begin{minipage}[b]{\linewidth}\raggedright
cosine similarity
\end{minipage} & \begin{minipage}[b]{\linewidth}\raggedright
definition
\end{minipage} \\
\midrule
\endhead
1 / \emph{catch-22} & 1.0 & \emph{an unpleasant situation from which you
cannot escape} \\
2 / \emph{life-or-death} & 0.3943 & \emph{extremely serious, especially
when there is a situation in which people might die} \\
3 / \emph{apples and oranges} & 0.3924 & \emph{two people or things are
completely different from each other} \\
4 / \emph{rocket science} & 0.38 & \emph{used to emphasize that
something is easy to do or understand} \\
5 / \emph{double-edged sword} & 0.3801 & \emph{to be something that has
both advantages and disadvantages} \\
\bottomrule
\end{longtable}

A python library, \texttt{idiom2vec}, was developed to train \& explore
Idiom2Vec models. The library offers a way of exploring the nearest
idioms (in terms of cosine similarity) to a given idiom that an
Idiom2Vec model has learned. An example of this with \emph{catch-22} as
the input is illustrated in \textbf{Table 11} above. Here, we see that
the model has learned the juxtaposing nature of \emph{catch-22}, and
thus infer that \emph{life-or-death} (live or die), \emph{apples and
oranges} (two different things) and \emph{double-edged sword} (pros and
cons) are similar in the semantics to \emph{catch-22}, purely from raw
data.

\hypertarget{reverse-search-idioms-idiomify}{%
\subsection{3.5. Reverse-search idioms
(Idiomify)}\label{reverse-search-idioms-idiomify}}

\begin{quote}
\begin{Shaded}
\begin{Highlighting}[]
\KeywordTok{def}\NormalTok{ phrase\_vector(}\VariableTok{self}\NormalTok{, phrase: }\BuiltInTok{str}\NormalTok{) }\OperatorTok{{-}\textgreater{}}\NormalTok{ np.array:}
    \CommentTok{\# 1. refine and tokenize the phrase}
\NormalTok{    tokens }\OperatorTok{=} \VariableTok{self}\NormalTok{.refine(phrase)}
    \CommentTok{\# 2. get available tokens}
\NormalTok{    avail\_tokens }\OperatorTok{=}\NormalTok{ [}
\NormalTok{        token}
        \ControlFlowTok{for}\NormalTok{ token }\KeywordTok{in}\NormalTok{ tokens}
        \ControlFlowTok{if} \VariableTok{self}\NormalTok{.idiom2vec\_wv.wv.key\_to\_index.get(token, }\VariableTok{None}\NormalTok{)}
\NormalTok{    ]}
    \CommentTok{\# 4. abort if nothing has been learnt}
    \ControlFlowTok{if} \BuiltInTok{len}\NormalTok{(avail\_tokens) }\OperatorTok{==} \DecValTok{0}\NormalTok{:}
        \ControlFlowTok{return} \VariableTok{None}
    \CommentTok{\# 3. vectorize the tokens}
\NormalTok{    token\_vectors }\OperatorTok{=}\NormalTok{ [}
        \VariableTok{self}\NormalTok{.idiom2vec\_wv.wv.get\_vector(token)}
        \ControlFlowTok{for}\NormalTok{ token }\KeywordTok{in}\NormalTok{ avail\_tokens}
\NormalTok{    ]}
    \CommentTok{\# 5. return their average}
    \ControlFlowTok{return}\NormalTok{ np.array(token\_vectors).mean(axis}\OperatorTok{=}\DecValTok{0}\NormalTok{)}
\end{Highlighting}
\end{Shaded}

Code 2: The Python implementation for vectorzing an input phrase
\end{quote}

We reverse-search idioms by averaging the Idiom2Vec embeddings of a
given phrase to get a phrase vector, then searching for the nearest
idiom neighbours to the phrase vector. \textbf{Code 2} above shows how
the phrase vector is obtained; First, the input phrase is refined and
tokenised. Any numbers, punctuations are removed, and each word in a
phrase is lemmatised. Next, we get only the tokens that the Idiom2Vec
has seen in the training. For example, if \texttt{tokens} were
\texttt{{[}excitedly,\ and,\ anxiously{]}} but \texttt{self.idiom2vec}
were Idiom2Vec V3, then only \texttt{{[}excitedly,\ anxiously{]}} would
be saved in \texttt{avail\_tokens} because V3 is trained with
non-stopwords data (as discussed in Section \textbf{3.4.}). If no tokens
are available, the method returns \texttt{None} to abort the process.
Otherwise, we vectorize the available tokens, the average of which is
then returned to give the phrase vector.

\begin{figure}
\centering
\includegraphics[width=4.79167in,height=\textheight]{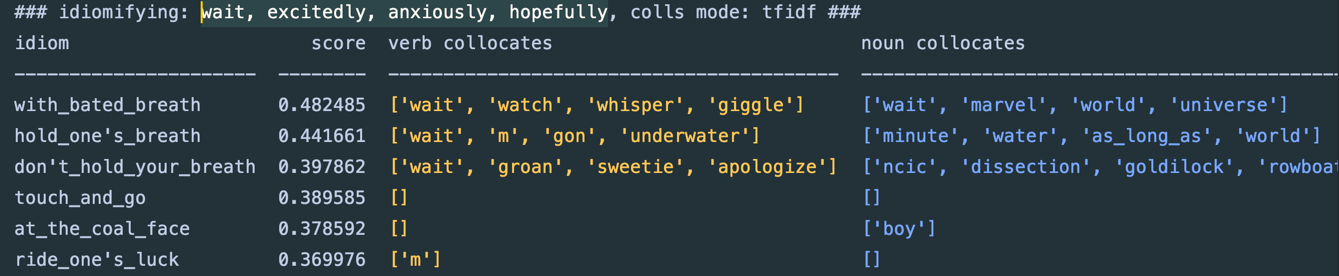}
\caption{The results of idiomifying \emph{wait, anxiously, excitedly,
hopefully}}
\end{figure}

\begin{figure}
\centering
\includegraphics[width=4.79167in,height=\textheight]{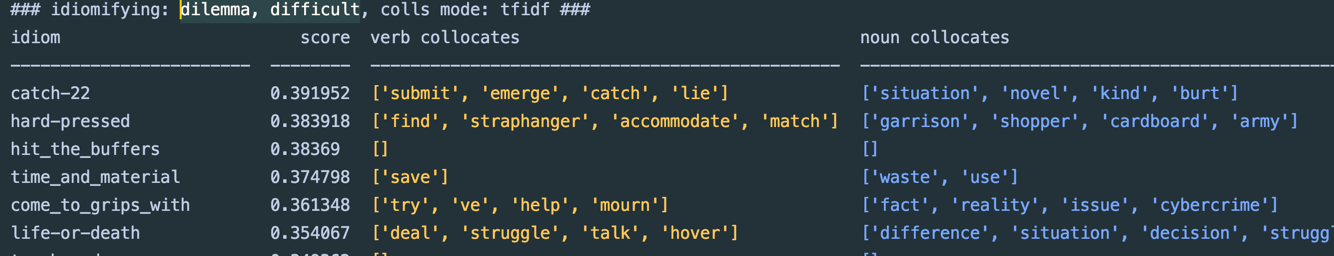}
\caption{The results of idiomifying \emph{difficult, dilemma}}
\end{figure}

A python library, \texttt{idiomify}, was developed to amalgamate
\texttt{idiom2vec} and \texttt{idiom2collocations} into Idiomify, a
collocation-supplemented reverse-dictionary of idioms. \textbf{Figure 8}
and \textbf{Figure 9} each illustrate an example usage of Idiomify. On
idiomifying \emph{wait, anxiously, excitedly and hopefully}, we get
\emph{with bated breath} as the most semantically similar idiom to the
set of words (\textbf{Figure 8}). Likewise, on idiomifying
\emph{difficult, dilemma}, we get \emph{catch-22} as the most relevant
(\textbf{Figure 9}). The results are supplemented with the verb, noun,
adjective and noun collocations of each idiom in the result. Although
the adjective and adverb collocations are truncated in the figures, they
are observable by horizontally dragging to the right on the results.

\pagebreak

\hypertarget{evaluation}{%
\section{4. Evaluation}\label{evaluation}}

\hypertarget{identifying-idioms-2}{%
\subsection{4.1. Identifying idioms}\label{identifying-idioms-2}}

\hypertarget{measures}{%
\subsubsection{4.1.1. Measures}\label{measures}}

We evaluate the flexibility of the idiom-matching rules by testing them
on an example of each variation case of idioms (total of six, discussed
in Section \textbf{2.1}). For this, a representative example for each of
case are chosen from \emph{SLIDE} (introduced in Section \textbf{3.1}).
We test if the matching rules can identify the idioms from their
exemplar use cases. For instance, as for testing for the alternatives
case, we test if the rules can identify \emph{add fuel to the fire} from
\emph{others in the media \textbf{threw gasoline on the fire} by blaming
farmers}, where \emph{throw gasoline on the fire} is an alternative form
to \emph{add fuel to the fire}. The results of the tests are presented
in \textbf{Table 12} and \textbf{Table 13} below.

\hypertarget{results-analysis}{%
\subsubsection{4.1.2. Results \& analysis}\label{results-analysis}}

\begin{longtable}[]{@{}
  >{\raggedright\arraybackslash}p{(\columnwidth - 4\tabcolsep) * \real{0.3333}}
  >{\raggedright\arraybackslash}p{(\columnwidth - 4\tabcolsep) * \real{0.3333}}
  >{\raggedright\arraybackslash}p{(\columnwidth - 4\tabcolsep) * \real{0.3333}}@{}}
\caption{The test results of the three positive cases: optional hyphen,
inflection and alternatives.}\tabularnewline
\toprule
\begin{minipage}[b]{\linewidth}\raggedright
case
\end{minipage} & \begin{minipage}[b]{\linewidth}\raggedright
sentence
\end{minipage} & \begin{minipage}[b]{\linewidth}\raggedright
filtered idiom lemma
\end{minipage} \\
\midrule
\endfirsthead
\toprule
\begin{minipage}[b]{\linewidth}\raggedright
case
\end{minipage} & \begin{minipage}[b]{\linewidth}\raggedright
sentence
\end{minipage} & \begin{minipage}[b]{\linewidth}\raggedright
filtered idiom lemma
\end{minipage} \\
\midrule
\endhead
optional hyphen(hyphenated) & \emph{in terms of rhyme, meter, and
\textbf{balls-out} swagger.} &
\texttt{{[}\textquotesingle{}balls-out\textquotesingle{}{]}} \\
optional hyphen(hyphen omitted) & \emph{in terms of rhyme, meter, and
\textbf{balls out} swagger.} &
\texttt{{[}\textquotesingle{}balls-out\textquotesingle{}{]}} \\
inflection(someone's) & \emph{they were \textbf{teaching me a lesson}
for daring to complain.} &
\texttt{{[}\textquotesingle{}teach\ someone\ a\ lesson\textquotesingle{}{]}} \\
inflection(one's) & \emph{Jo is a playwright who has always \textbf{been
ahead of her time}} &
\texttt{{[}\textquotesingle{}ahead\ of\ one\textquotesingle{}s\ time\textquotesingle{}{]}} \\
alternatives(original) & \emph{others in the media have \textbf{added
fuel to the fire} by blaming farmers} &
\texttt{{[}\textquotesingle{}add\ fuel\ to\ the\ fire\textquotesingle{}{]}} \\
alternatives(1) & \emph{others in the media have \textbf{added fuel to
the flame} by blaming farmers} &
\texttt{{[}\textquotesingle{}add\ fuel\ to\ the\ fire\textquotesingle{}{]}} \\
alternatives(2) & \emph{others in the media have \textbf{poured gasoline
on the fire} by blaming farmers} &
\texttt{{[}\textquotesingle{}add\ fuel\ to\ the\ fire\textquotesingle{}{]}} \\
alternatives(3) & \emph{others in the media have \textbf{threw gasoline
on the fire} by blaming farmers} &
\texttt{{[}\textquotesingle{}add\ fuel\ to\ the\ fire\textquotesingle{}{]}} \\
alternatives(4) & \emph{others in the media have \textbf{threw gas on
the fire} by blaming farmers} &
\texttt{{[}\textquotesingle{}add\ fuel\ to\ the\ fire\textquotesingle{}{]}} \\
\bottomrule
\end{longtable}

The matching rules are able to identify idioms with an optional hyphen,
those with inflecting words and those in their alternative forms. As can
be seen from the first two rows of \textbf{Table 12}, the patterns can
correctly identify \emph{balls-out} regardless of whether the hyphen is
omitted or not. The patterns can also handle the case of inflections;
they can identify \emph{teach someone a lesson} from \emph{teaching me a
lesson} (personal pronoun inflection, the third row) and \emph{ahead of
one's time} from \emph{ahead of her time} (possesive pronoun inflection,
the fourth row). Lastly, we see that the patterns can identify \emph{add
fuel to the fire} from its four alternative forms: \emph{add fuel to the
flame}, \emph{pour gasoline on the fire}, \emph{throw gasoline on the
fire}, \emph{throw gas on the fire}.

\begin{longtable}[]{@{}
  >{\raggedright\arraybackslash}p{(\columnwidth - 4\tabcolsep) * \real{0.3333}}
  >{\raggedright\arraybackslash}p{(\columnwidth - 4\tabcolsep) * \real{0.3333}}
  >{\raggedright\arraybackslash}p{(\columnwidth - 4\tabcolsep) * \real{0.3333}}@{}}
\caption{The test results of the three negative cases: modification,
open slot and passivisation.}\tabularnewline
\toprule
\begin{minipage}[b]{\linewidth}\raggedright
case
\end{minipage} & \begin{minipage}[b]{\linewidth}\raggedright
sentence
\end{minipage} & \begin{minipage}[b]{\linewidth}\raggedright
filtered idiom lemma
\end{minipage} \\
\midrule
\endfirsthead
\toprule
\begin{minipage}[b]{\linewidth}\raggedright
case
\end{minipage} & \begin{minipage}[b]{\linewidth}\raggedright
sentence
\end{minipage} & \begin{minipage}[b]{\linewidth}\raggedright
filtered idiom lemma
\end{minipage} \\
\midrule
\endhead
\emph{modification} & \emph{He \textbf{grasped at straws}} - \emph{He
\textbf{grasped desperately at the floating straw.}} &
\texttt{{[}\textquotesingle{}grasp\ at\ straws\textquotesingle{}{]}} -
\texttt{{[}{]}} \\
\emph{passivisation} & \emph{And with him gone, they \textbf{opened the
floodgates.}} - \emph{And with him gone, \textbf{the floodgates were
opened.}} & \texttt{{[}\textquotesingle{}open\ the\ floodgamtes{]}} -
\texttt{{[}{]}} \\
\emph{open slot} & \emph{They preferred to persist in \textbf{keeping
them at arm's length}.} - \emph{They preferred to persist in
\textbf{keeping both Germans and Russians at arm's length.}} &
\texttt{{[}\textquotesingle{}keep\ someone\ at\ arm\textquotesingle{}s\ length\textquotesingle{}{]}}
-
\texttt{{[}\textquotesingle{}at\ arm\textquotesingle{}s\ length\textquotesingle{}{]}} \\
\bottomrule
\end{longtable}

Despite the three positive cases, the idiom-matching rules are unable to
cover the case of modification, passivisation and open slot. Although
idioms could be modified with modals, be passivised, and their open
slots, i.e.~\emph{one's} and \emph{someone's}, could be replaced with a
set of words (as discussed in Section \textbf{2.1}), the rules do not
take into account all the three cases. For instance, as can be seen in
\textbf{Table 13}, The rules can no longer identify \emph{grasp at
straws} when it is modified to \emph{grasped desperately at the floating
straw} (the first row). Likewise, they can no longer identify \emph{open
the floodgates} when it is written in a passive tense, as in \emph{the
floodgates were opened}(the second row). While they can identify
\emph{at arm's length} from \emph{\ldots keeping both Germans and
Russians at arm's length}, the identified idiom is nonetheless not the
correct idiom at use (the third row); It should identify \emph{keep
someone at arm's length}, not just \emph{at arm's lenth}.

\hypertarget{discussion}{%
\subsubsection{4.1.3. Discussion}\label{discussion}}

\begin{longtable}[]{@{}
  >{\raggedright\arraybackslash}p{(\columnwidth - 4\tabcolsep) * \real{0.3333}}
  >{\raggedright\arraybackslash}p{(\columnwidth - 4\tabcolsep) * \real{0.3333}}
  >{\raggedright\arraybackslash}p{(\columnwidth - 4\tabcolsep) * \real{0.3333}}@{}}
\caption{A summary of ElasticSearch query generation for retrieving
idioms (Adapted from: Hughes et al., 2021)}\tabularnewline
\toprule
\begin{minipage}[b]{\linewidth}\raggedright
variation
\end{minipage} & \begin{minipage}[b]{\linewidth}\raggedright
solution
\end{minipage} & \begin{minipage}[b]{\linewidth}\raggedright
Example ElasticSearch query
\end{minipage} \\
\midrule
\endfirsthead
\toprule
\begin{minipage}[b]{\linewidth}\raggedright
variation
\end{minipage} & \begin{minipage}[b]{\linewidth}\raggedright
solution
\end{minipage} & \begin{minipage}[b]{\linewidth}\raggedright
Example ElasticSearch query
\end{minipage} \\
\midrule
\endhead
\emph{modification} & slop &
\texttt{\{“query”:\ “call\ someone’s\ bluff”,\ “slop”:\ 4\}} \\
\emph{open slot} & wildcard + slop &
\texttt{\{“query”:\ “call\ *\ bluff”,\ “slop”:\ 5\}} \\
\emph{passivisation} with \emph{modification} & reordering + slop &
\texttt{\{“query”:\ “someone’s\ bluff\ *\ call”,\ “slop”:\ 5\}} \\
\emph{passivisation} with an \emph{open slot} & reordering + wildcard +
slop & \texttt{\{“query”:\ “*\ bluff\ *\ call”,\ “slop”:\ 6\}} \\
\bottomrule
\end{longtable}

We could improve the idiom-matching rules if we incorporate slop,
wildcard and reordering techniques into the patterns. This could have
them cover modification, open slot and passivisation cases.
\textbf{Table 14} illustrates how Hughes et al.~(2021) achieve this,
with \emph{call someone's bluff} as an example. Although they use
ElasticSearch to implement their solutions, we could still adopt the
techniques to address the three negative examples introduced in
\textbf{Table 13} above. For instance, we could match \emph{grasped
desperately at the floating straw} with ``grasp (slop) at (slop)
straws'' as ``slop'' would allow rooms for any words in between the
constituent words of an idiom, thereby addressing the modification case.
We could also match \emph{keeping both Germans and Russians at arm's
length} with ``keeping (slop) * (slop) at arm's length'', as the
wildcard would allow for the open slot to be substituted with any word.
We should also be able to match \emph{the floodgates were opened} with
``* the floodgates * open'' as the position of the main verb is now
reordered to address passivisation.

\hypertarget{modeling-collocations-of-idioms-1}{%
\subsection{4.2. Modeling Collocations of
Idioms}\label{modeling-collocations-of-idioms-1}}

\hypertarget{measures-1}{%
\subsubsection{4.2.1. Measures}\label{measures-1}}

\begin{longtable}[]{@{}
  >{\raggedright\arraybackslash}p{(\columnwidth - 4\tabcolsep) * \real{0.3333}}
  >{\raggedright\arraybackslash}p{(\columnwidth - 4\tabcolsep) * \real{0.3333}}
  >{\raggedright\arraybackslash}p{(\columnwidth - 4\tabcolsep) * \real{0.3333}}@{}}
\caption{Examples of idioms from group A(1-2), B(77-78), C(195-199)
D(555-567) and E(23115-142905) with their definitions. There are 20
idioms in each group, 100 idioms in total, but here we show one example
from each frequency group. This test dataset is used to evaluate both
the collocation of idioms, and the reverse-dictionary of idioms (as will
be discussed in Section \textbf{4.3.1}).}\tabularnewline
\toprule
\begin{minipage}[b]{\linewidth}\raggedright
idiom
\end{minipage} & \begin{minipage}[b]{\linewidth}\raggedright
frequency/group
\end{minipage} & \begin{minipage}[b]{\linewidth}\raggedright
definition
\end{minipage} \\
\midrule
\endfirsthead
\toprule
\begin{minipage}[b]{\linewidth}\raggedright
idiom
\end{minipage} & \begin{minipage}[b]{\linewidth}\raggedright
frequency/group
\end{minipage} & \begin{minipage}[b]{\linewidth}\raggedright
definition
\end{minipage} \\
\midrule
\endhead
\emph{from a to z} & 2/A & over the entire range \\
\emph{spectator sport} & 77/B & something that people watch other people
do without becoming involved themselves \\
\emph{best of both worlds} & 195/C & all the advantages of two different
situations and none of the disadvantages \\
\emph{have one's hands full} & 555/D & To be busy or thoroughly
preoccupied. \\
\emph{tell you what} & 23115/E & used to introduce a suggestion \\
\bottomrule
\end{longtable}

We evaluate the quality of the collocations of idioms by checking if
they agree with the use cases of idioms as stated in dictionaries.
Inspecting the collocations of idioms manually is costly. There are no
ready-made test set for collocations of idioms either. However,
compilers of dictionaries tend to put the most collocative use cases
when they compile example sentences for words or idioms. Therefore, we
can assume that the words used in dictionary-stated use cases of idioms
are collocates of the idioms, and thus use them as the ground truth to
evaluate the extracted collocations with. We evaluate in this way the
top 5 collocations of the five idioms listed in \textbf{Table 15} above.
For instance, we collect the example sentences for \emph{have one's
hands full} from Oxford, Cambridge and Merriam-Webster dictionaries
(\textbf{Table 16}), with which we evaluate the top 5 extracted
collocations of \emph{have one's hands full} (\textbf{Table 17}). The
collocates that show agreement with the example sentences are bolded
(e.g.~the noun collocate \textbf{\emph{at the moment}} agrees with
\emph{\ldots{} I have my hands full \textbf{right now}}). The results
for other idioms are included in Appendix at the end of the report:
\emph{from a to z} (Appendix A1 \& A2), \emph{spectator sport} (Appendix
B1 and B2), \emph{best of both worlds} (Appendix C1 and C2), and
\emph{tell you what} (Appendix D1 and D2).

\hypertarget{results-analysis-1}{%
\subsubsection{4.2.2. Results \& analysis}\label{results-analysis-1}}

\begin{longtable}[]{@{}
  >{\raggedright\arraybackslash}p{(\columnwidth - 4\tabcolsep) * \real{0.3333}}
  >{\raggedright\arraybackslash}p{(\columnwidth - 4\tabcolsep) * \real{0.3333}}
  >{\raggedright\arraybackslash}p{(\columnwidth - 4\tabcolsep) * \real{0.3333}}@{}}
\caption{The representative use cases of \emph{have one's hands
full}(freq 555, group D).}\tabularnewline
\toprule
\begin{minipage}[b]{\linewidth}\raggedright
oxford
\end{minipage} & \begin{minipage}[b]{\linewidth}\raggedright
cambridge
\end{minipage} & \begin{minipage}[b]{\linewidth}\raggedright
merriam webster
\end{minipage} \\
\midrule
\endfirsthead
\toprule
\begin{minipage}[b]{\linewidth}\raggedright
oxford
\end{minipage} & \begin{minipage}[b]{\linewidth}\raggedright
cambridge
\end{minipage} & \begin{minipage}[b]{\linewidth}\raggedright
merriam webster
\end{minipage} \\
\midrule
\endhead
\emph{She \textbf{certainly} has her hands \textbf{full with four kids}
in the house} & \emph{I'm sorry I can't help you -- I have my hands full
\textbf{right now}} & \emph{She'll have her hands full \textbf{with the
new baby.}} \\
\bottomrule
\end{longtable}

\begin{longtable}[]{@{}
  >{\raggedright\arraybackslash}p{(\columnwidth - 8\tabcolsep) * \real{0.2000}}
  >{\raggedright\arraybackslash}p{(\columnwidth - 8\tabcolsep) * \real{0.2000}}
  >{\raggedright\arraybackslash}p{(\columnwidth - 8\tabcolsep) * \real{0.2000}}
  >{\raggedright\arraybackslash}p{(\columnwidth - 8\tabcolsep) * \real{0.2000}}
  >{\raggedright\arraybackslash}p{(\columnwidth - 8\tabcolsep) * \real{0.2000}}@{}}
\caption{Top 5 collocations and use cases for \emph{have one's hands
full}(freq 555, group D).}\tabularnewline
\toprule
\begin{minipage}[b]{\linewidth}\raggedright
model
\end{minipage} & \begin{minipage}[b]{\linewidth}\raggedright
verb
\end{minipage} & \begin{minipage}[b]{\linewidth}\raggedright
noun
\end{minipage} & \begin{minipage}[b]{\linewidth}\raggedright
adj
\end{minipage} & \begin{minipage}[b]{\linewidth}\raggedright
adv
\end{minipage} \\
\midrule
\endfirsthead
\toprule
\begin{minipage}[b]{\linewidth}\raggedright
model
\end{minipage} & \begin{minipage}[b]{\linewidth}\raggedright
verb
\end{minipage} & \begin{minipage}[b]{\linewidth}\raggedright
noun
\end{minipage} & \begin{minipage}[b]{\linewidth}\raggedright
adj
\end{minipage} & \begin{minipage}[b]{\linewidth}\raggedright
adv
\end{minipage} \\
\midrule
\endhead
TF & gon, try, ve, think, know & today, gon, guy, \textbf{mother},
police & sure, \textbf{new}, private, separate, dear & right,
\textbf{certainly}, kind, kinda, clearly \\
TFIDF & gon, ve, try, deal, doubtless & gon, \textbf{at the moment},
police, \textbf{nurse}, journalist & sure, uterine, overloaded,
\textbf{new}, liable & kinda, \textbf{certainly}, right, kind,
clearly \\
PMI & investigate, ve, search, suspect, deal, gon & \textbf{at the
moment}, gon,\textbf{nurse}, journalist, guard, bunch & sure,
\textbf{new} & kinda, \textbf{certainly}, clearly, kind, obviously,
right, \\
\bottomrule
\end{longtable}

PMI \& TF-IDF generally rank the collocations better than TF does. One
example that shows this is the collocations extracted for \emph{have
one's hands full}, which are represented in \textbf{Table 17}. From the
Cambridge's example of the idiom (\textbf{Table 16}), we know that the
idiom collocates with the word \emph{now}, which is in agreement with
the verb collocates extracted with TFIDF \& PMI (since they both ranked
\emph{at the moment} high). We however do not see \emph{at the moment}
in the verb collocates extracted with TF. This is reassuring, as TF can
only measure ``frequently co-occur'' part of the definition of
collocation, while PMI and TF-IDF can take into account ``uniquely
co-occur'' part of it as well.

It is nonetheless hard to discern whether PMI is more suitable than
TF-IDF or vice versa. Although this may be partly due to a lack of
reliable test data, the two metrics are hardly different in the way they
have ranked the ground truth collocates. For example, both PMI and
TF-IDF ranked \emph{certainly} the same for the adjective collocates of
\emph{have one's hands full}. Though PMI ranked \emph{new} higher than
TF-IDF did for the adjective collocates, the marginal difference between
PMI and TF-IDF is well observable also in the collocations extracted for
the idioms in other frequency groups as well (e.g.~Appendix \textbf{1A},
\textbf{2A}, \textbf{3A}.).

\hypertarget{discussion-1}{%
\subsubsection{4.2.3. Discussion}\label{discussion-1}}

\begin{longtable}[]{@{}
  >{\raggedright\arraybackslash}p{(\columnwidth - 4\tabcolsep) * \real{0.3333}}
  >{\raggedright\arraybackslash}p{(\columnwidth - 4\tabcolsep) * \real{0.3333}}
  >{\raggedright\arraybackslash}p{(\columnwidth - 4\tabcolsep) * \real{0.3333}}@{}}
\caption{The representative use cases of \emph{from a to z}(freq 1,
group A).}\tabularnewline
\toprule
\begin{minipage}[b]{\linewidth}\raggedright
Oxford
\end{minipage} & \begin{minipage}[b]{\linewidth}\raggedright
Cambridge
\end{minipage} & \begin{minipage}[b]{\linewidth}\raggedright
Merriam-Webster
\end{minipage} \\
\midrule
\endfirsthead
\toprule
\begin{minipage}[b]{\linewidth}\raggedright
Oxford
\end{minipage} & \begin{minipage}[b]{\linewidth}\raggedright
Cambridge
\end{minipage} & \begin{minipage}[b]{\linewidth}\raggedright
Merriam-Webster
\end{minipage} \\
\midrule
\endhead
\emph{He knew his subject from A to Z} & \emph{This book tells the story
of her life from A to Z.} & \emph{The book is titled ``Home Repairs From
A to Z.''} \\
\bottomrule
\end{longtable}

\begin{longtable}[]{@{}lllll@{}}
\caption{Top 5 collocations and use cases for \emph{from a to z}(freq 1,
group A).}\tabularnewline
\toprule
model & verb & noun & adj & adv \\
\midrule
\endfirsthead
\toprule
model & verb & noun & adj & adv \\
\midrule
\endhead
tf & scan & - & - & - \\
tfidf & scan & - & - & - \\
pmi & - & - & - & - \\
\bottomrule
\end{longtable}

We should collect more data for the idioms in frequency group A. As can
be seen from \textbf{Table 18} above, the three example sentences of
\emph{from a to z} clearly indicate that the noun \emph{book} collocates
with \emph{from a to z}. However, there were only one instance found for
\emph{from a to z}. It is for this reason we were unable to discover the
obvious collocate \emph{book}, let alone any noun, adjective or adverb
collocates (\textbf{Table 19}).

If the problem is that we are torn between PMI and TF-IDF, machine
learning could help us get the best of both worlds. As discussed
earlier, it is hard to tell which of the two metrics are better than the
other. Hence, it is not ideal to pick a single measure for modeling
collocations. However, it is impossible for humans to determine the
optimal preference for PMI and TF-IDF. In his paper, \emph{Machine
Learning For Collocation Identification}, Yang argues (2003) that this
is a ``niche for machine learning''. That is, we can take advantage of
multiple measures of collocations if we can learn the optimal preference
for each metric from data. Yang does exactly that; he uses the two-word
pair entries in \emph{WordNet} as the ground-truth for collocations,
from which a Decision Tree model learns the optimal preferences for TF,
T-test score, PMI, dice coefficient, log-likelihood ratio, and I-score.
This can be directly transferable to our case because the notion of
collocations is the same for words and idioms. We could try replicating
the same approach, and use the class weights of PMI and TF-IDF learned
by a Decision Tree model to retrieve the optimal collocations of idioms.

\hypertarget{reverse-searching-idioms-idiomify}{%
\subsection{4.3. Reverse-searching idioms
(Idiomify)}\label{reverse-searching-idioms-idiomify}}

\hypertarget{measures-2}{%
\subsubsection{4.3.1. Measures}\label{measures-2}}

We evaluate the performance of Idiomify (the reverse dictionary) with
median ranks as the evaluation metric. That is, for each idiom in
\textbf{Table 15} (from Section \textbf{4.2.1}), we reverse-search them
on Idiomify with their definitions (e.g.~search ``To be busy or
thoroughly preoccupied'' to find the rank of \emph{have one's hands
full}), sum of their ranks in the list of suggested idioms that Idiomify
suggests, and compute their median. While there are many other metrics
for evaluating a search engine, such as MMR, NDCG and MAP (cite
professor Nenadic's slide - which slide was this?), we use median ranks
here because it is what Hill et al.~used (2016) to evaluate their
reverse-dictionary. By having the same metric as theirs, we can have a
concrete reference with which we can compare the performance of
Idiomify. The result of this is represented in \textbf{Table 20} below.

\hypertarget{results-analysis-2}{%
\subsubsection{4.3.2. Results \& Analysis}\label{results-analysis-2}}

\begin{longtable}[]{@{}llll@{}}
\caption{The comparison of Idiomify with the models devised by Hill et
al.~(2016). Idiomify was tested on \textbf{Table 15}. all the other
models are tested on an internal test set that Hill et al.~built
themselves.}\tabularnewline
\toprule
model & method & median rank & variance \\
\midrule
\endfirsthead
\toprule
model & method & median rank & variance \\
\midrule
\endhead
Idiomify (Idiom2Vec) & avg & 128 & 4310 \\
Word2Vec (Hill et al., 2016) & add & 923 & 163 \\
Word2Vec (Hill et al., 2016) & mul & 1000 & 10 \\
OneLook & - & 0 & 67 \\
RNN cosine (Hill et al., 2016) & - & 12 & 103 \\
\bottomrule
\end{longtable}

Idiomify sets itself as a reasonable baseline for the reverse dictionary
of idioms. As can be seen in \textbf{Table 20} above, Idiomify seems to
perform better than all the Word2Vec model that Hill et al.~devised
(2016). although this is encouraging, it should be noted that this is
not a fair comparison; the test set and search space is vastly different
(2714 idioms vs.~all the words), and the large variance might be marring
the median. When compared with the RNN cosine model, it is also by no
means an outstanding reverse-dictionary of words. However, when it comes
to that of idioms, it could be argued that Idiomify is clearly a good
baseline to improve upon, as it is the only one of its kind at the
moment.

\hypertarget{discussion-2}{%
\subsubsection{4.3.3. Discussion}\label{discussion-2}}

\begin{longtable}[]{@{}lllll@{}}
\caption{Performance of idiomify by different versions of
Idiom2Vec.}\tabularnewline
\toprule
Idiom2Vec & corpora & stopwords & median rank & \\
\midrule
\endfirsthead
\toprule
Idiom2Vec & corpora & stopwords & median rank & \\
\midrule
\endhead
V1 & COCA (spok) & not removed & 276.5 & \\
V2 & COCA (spok) + Opensubtitles & not removed & 128.0 & \\
V3 & COCA (spok) + Opensubtitles & removed & 156.0 & \\
\bottomrule
\end{longtable}

In future work, stopwords should not be excluded in building a
reverse-dictionary. This is because, as shown in \textbf{Table 20}
above, we see that median rank of Idiomify increases by 28 when we
switch from Idiom2Vec V2 to V3, where V3 is trained on non-stopwords
corpus. That is, the performance of Idiomify drops when we exclude
stopwords from the training data. It is speculated that this is because
the stopwords do hold semantic significance when they are used to
defining idioms or words. For example, \emph{all the advantages of two
different situations and none of the disadvantages} clearly means
\emph{best of both worlds}, but it is less clear if \emph{advantages,
two, different, situation, disadvantages} would mean precisely the same
idiom. Despite being stopwords, the meaning of \emph{all} and
\emph{none} are just as integral to the meaning of \emph{best of both
worlds} as the non-stopwords. Hence, as far as reverse-dictionaries are
concerned, stopwords are useful and therefore should not be removed from
any data.

\begin{figure}
\centering
\includegraphics[width=3.64583in,height=\textheight]{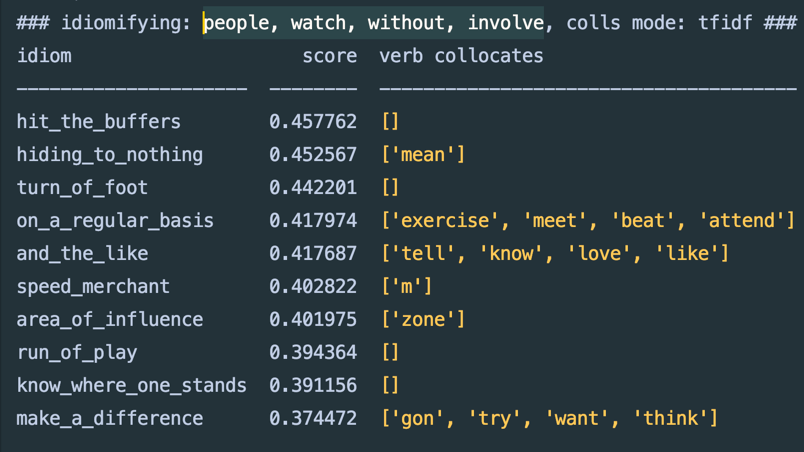}
\caption{the result of Idiomifying \emph{people, watch, without,
involve}.}
\end{figure}

We could further improve Idiomify by taking a mixture of inverted index
and distributional semantics approach. Although a purely distributional
semantics approach allows for semantic search, they require a large
amount of quality data to work properly. Without enough data, it would
perform even worse than a simple inverted index. \textbf{Figure 10} well
illustrates this problem. It attempts to idiomify a set of keywords from
the definition of \emph{spectator sport}, but \emph{spectator sport} is
not even on the list. This is as it should be, as Idiom2Vec V2 was
trained on only 75 instances of the idiom. In contrast to this, we would
only need a dictionary of idioms if we were to take an inverted index
approach. That way, reverse searching idioms with their exact
definitions, as is the case in \textbf{Figure 10}, would be trivial.
However, semantic search would not be possible if we only use an
inverted index. All things considered, then, the best approach would be
to use both. That is, if we could have Idiomify first search on a
pre-populated def-to-idiom inverted index, and later rank the retrieved
idioms with respect to their similarity to Idiom2Vec-averaged phrase
vector, We could conceivably improve both the accuracy and flexibility
of Idiomify.

\pagebreak

\hypertarget{conclusion}{%
\section{5. Conclusion}\label{conclusion}}

The aim of idiomify is to build a collocation-supplemented reverse
dictionary of idioms for the non-native learners of English. We aim to
do so because the reverse dictionary could help the non-natives explore
idioms on demand, and the collocations could also guide them on using
idioms more adequately. The cornerstone of the project is a reliable way
of mining idioms from corpora, which is however a challenge because
idioms extensively vary in forms. We tackle this by automatically
deriving matching rules from their base forms. We use Point-wise Mutual
Inclusion (PMI), Term Frequency - Inverse Document Frequency (TF-IDF) to
model collocations, since both of them are popular metric for pairwise
significance. We also try Term Frequency (TF) as the baseline model. As
for implementing the reverse-dictionary, three approaches could be
taken: inverted index, graphs and distributional semantics. We choose to
take the last approach and implement the reverse dictionary with
Word2Vec, because it is the most flexible approach of all and Word2Vec
is a simple yet strong baseline. Evaluating the methods has revealed
rooms for improvement. We learn that we can better identify idioms with
the help of slop, wildcard and reordering techniques. We also learn that
we can get the best of both PMI and TF-IDF if we use machine learning to
find the sweet spot. Lastly, We learn that Idiomify could be further
improved with a mixture of inverted index and distributional semantics
approach. The limits aside, the proposed methods are feasible, and their
benefits to the non-natives are apparent, which therefore can be used to
aid the non-natives in acquiring English idioms.

\pagebreak

\hypertarget{references}{%
\subsection{6. References}\label{references}}

Arora, S. Liang, Y. Ma, T. (2017). \emph{A Simple but tough-to-beat
baseline for sentence embeddings}

Beeferman, D. (2003). \emph{OneLook reverse dictionary}. Available at:
\url{http://onelook.com/reversedictionary.shtml}.

Caro, R. Edith, E. (2019). \emph{The Advantages and Importance of
Learning and Using Idioms in English}.

Church, K. Hanks, P. (1989). \emph{Word Association Norms, Mutual
Information, and Lexicography}

Davies, M. (2008). \emph{The Corpus of Contemporary American English}
(COCA). Available online at \url{https://www.english-corpora.org/coca/}.

Ganesan, K. (2020). \emph{What are Stop Words?}. Available at:
\url{https://kavita-ganesan.com/what-are-stop-words/\#.YKfXMy0Rr_Q}

Granger, S. (1998). \emph{Prefabricated patterns in advanced EFL
writing: collocations and formulae}. In: \emph{Phraseology: Theory,
Analysis and Applications}. Clarendon Press: Oxford. 145-160.

Gormley, C. Tony, Z. (2015). \emph{Elasticsearch: The Definitive Guide;
O'Reilly Media}. In: Sebastopol, CA, USA.

Hill, F. Cho, K. Korhonen, A. Bengio, Y. (2016). \emph{Learning to
Understand Phrases by Embedding the Dictionary}

Honnibal, M. Montani, I., (2017). \emph{spaCy 2: Natural language
understanding with Bloom embeddings, convolutional neural networks and
incremental parsing}.

Hughes, C. Filimonov, Maxim. Wray, Alison. Spasi´c, Irena. (2021).
\emph{Leaving No Stone Unturned: Flexible Retrieval of Idiomatic
Expressions from a Large Text Corpus}.

Ingebrigsten, M. (2013), \emph{Indexing for Beginners, Part 3}.
Available at:
\url{https://www.elastic.co/blog/found-indexing-for-beginners-part3}

Jochim, C. et al.~(2018). \emph{SLIDE: a Sentiment Lexicon of Common
Idioms}. Proceedings of the Eleventh International Conference on
Language Resources and Evaluation (LREC'2018). European Languages
Resources Association (ELRA)

Manning, D., et al.~(2008). \emph{Tf-idf weighting}. Viewed on 23th of
February 2021. Available at:
\url{https://nlp.stanford.edu/IR-book/html/htmledition/tf-idf-weighting-1.html}

Miller, G. (1995). \emph{WordNet: A Lexical Database for English}.
Communications of the ACM Vol. 38, No.~11: 39-41.

\emph{Oxford Advanced Learner's Dictionary of Current English}. (1995).
Oxford University Press

\emph{Oxford Collocations Dictionary for Students of English}. (2002).
Oxford University Press

Rehurek, R. Sojka, P. (2011). \emph{Gensim--python framework for vector
space modelling.} NLP Centre, Faculty of Informatics, Masaryk
University, Brno, Czech Republic, 3(2).

Sierra, G. (2000). \emph{The onomasiological dictionary: a gap in
lexicography}. In Proceedings f the Ninth EURALEX International

Tiedmann, J. (2012). \emph{Parallel Data, Tools and Interfaces in OPUS}.
Proceedings of the Eight International Conference on Language Resources
and Evaluation (LREC'2012). European Language Resources Association
(ELRA)

Thorat, S. Choudhari, V. (2016). \emph{Implementing a Reverse
Dictionary, based on word definitions, using a Node-Graph Architecture}.

Thyrab, R. (2016). \emph{The Necessity of idiomatic expressions to
English Language learners}

Yang, S. (2003). \emph{machine learning for collocation identification}

\pagebreak

\hypertarget{appendix}{%
\section{7. Appendix}\label{appendix}}

\begin{longtable}[]{@{}
  >{\raggedright\arraybackslash}p{(\columnwidth - 8\tabcolsep) * \real{0.2000}}
  >{\raggedright\arraybackslash}p{(\columnwidth - 8\tabcolsep) * \real{0.2000}}
  >{\raggedright\arraybackslash}p{(\columnwidth - 8\tabcolsep) * \real{0.2000}}
  >{\raggedright\arraybackslash}p{(\columnwidth - 8\tabcolsep) * \real{0.2000}}
  >{\raggedright\arraybackslash}p{(\columnwidth - 8\tabcolsep) * \real{0.2000}}@{}}
\toprule
\begin{minipage}[b]{\linewidth}\raggedright
model
\end{minipage} & \begin{minipage}[b]{\linewidth}\raggedright
verb
\end{minipage} & \begin{minipage}[b]{\linewidth}\raggedright
noun
\end{minipage} & \begin{minipage}[b]{\linewidth}\raggedright
adj
\end{minipage} & \begin{minipage}[b]{\linewidth}\raggedright
adv
\end{minipage} \\
\midrule
\endhead
tf & grow, know, fledge, die, mean & number, suit, form, courtesy, today
& great, \textbf{popular}, complete, big, large & exactly, especially,
fast, kind, long \\
tfidf & moneymake, fledge, grow, recommend, die & tutoring, form,
courtesy, suit, spectacle & \textbf{popular}, anlocal, great, dry,
mathematical & exactly, generally, especially, fast, definitely \\
pmi & grow, know & suit, form, number & \textbf{popular}, great &
exactly \\
\bottomrule
\end{longtable}

Appendix 1A. Top 5 collocations and use cases for \emph{spectator
sport}(freq 77, group B).

\begin{longtable}[]{@{}
  >{\raggedright\arraybackslash}p{(\columnwidth - 4\tabcolsep) * \real{0.3333}}
  >{\raggedright\arraybackslash}p{(\columnwidth - 4\tabcolsep) * \real{0.3333}}
  >{\raggedright\arraybackslash}p{(\columnwidth - 4\tabcolsep) * \real{0.3333}}@{}}
\toprule
\begin{minipage}[b]{\linewidth}\raggedright
oxford
\end{minipage} & \begin{minipage}[b]{\linewidth}\raggedright
cambridge
\end{minipage} & \begin{minipage}[b]{\linewidth}\raggedright
merriam webster
\end{minipage} \\
\midrule
\endhead
\emph{What is the country's most \textbf{popular} spectator sport?} &
\emph{Football is certainly the biggest spectator sport in Britain.} &
\emph{For many, politics has become a spectator sport.''} \\
\bottomrule
\end{longtable}

Appendix 1B. The representative use cases of \emph{spectator sport}(freq
77, group B).

\begin{longtable}[]{@{}
  >{\raggedright\arraybackslash}p{(\columnwidth - 8\tabcolsep) * \real{0.2000}}
  >{\raggedright\arraybackslash}p{(\columnwidth - 8\tabcolsep) * \real{0.2000}}
  >{\raggedright\arraybackslash}p{(\columnwidth - 8\tabcolsep) * \real{0.2000}}
  >{\raggedright\arraybackslash}p{(\columnwidth - 8\tabcolsep) * \real{0.2000}}
  >{\raggedright\arraybackslash}p{(\columnwidth - 8\tabcolsep) * \real{0.2000}}@{}}
\toprule
\begin{minipage}[b]{\linewidth}\raggedright
model
\end{minipage} & \begin{minipage}[b]{\linewidth}\raggedright
verb
\end{minipage} & \begin{minipage}[b]{\linewidth}\raggedright
noun
\end{minipage} & \begin{minipage}[b]{\linewidth}\raggedright
adj
\end{minipage} & \begin{minipage}[b]{\linewidth}\raggedright
adv
\end{minipage} \\
\midrule
\endhead
tf & want, \textbf{chill}, laugh, let, m & way, milk, combination,
future, pajama & old, salty, fashioned, \textbf{wonderful}, able &
right, kind, inside \\
tfidf & \textbf{chill}, laugh, embody, \textbf{gettin}, want &
combination, milk, pajama, future, \textbf{city} & salty, fashioned,
\textbf{wonderful}, black, able & inside, kind, right \\
pmi & \textbf{chill}, laugh, want, let, m & way & - & - \\
\bottomrule
\end{longtable}

Appendix 2A. Top 5 collocations and use cases for \emph{best of both
worlds}(freq 195, group C).

\begin{longtable}[]{@{}
  >{\raggedright\arraybackslash}p{(\columnwidth - 4\tabcolsep) * \real{0.3333}}
  >{\raggedright\arraybackslash}p{(\columnwidth - 4\tabcolsep) * \real{0.3333}}
  >{\raggedright\arraybackslash}p{(\columnwidth - 4\tabcolsep) * \real{0.3333}}@{}}
\toprule
\begin{minipage}[b]{\linewidth}\raggedright
oxford
\end{minipage} & \begin{minipage}[b]{\linewidth}\raggedright
cambridge
\end{minipage} & \begin{minipage}[b]{\linewidth}\raggedright
merriam webster
\end{minipage} \\
\midrule
\endhead
\emph{If you \textbf{enjoy} the coast and the country, you'll
\textbf{get} the best of both worlds on this walk} & \emph{She works in
the \textbf{city} and lives in the country, so she \textbf{gets} the
best of both worlds} & \emph{I have the best of both worlds---a
\textbf{wonderful} family and a great job.} \\
\bottomrule
\end{longtable}

Appendix 2B. The representative use cases of \emph{best of both
worlds}(freq 195, group C).

\begin{longtable}[]{@{}
  >{\raggedright\arraybackslash}p{(\columnwidth - 8\tabcolsep) * \real{0.2000}}
  >{\raggedright\arraybackslash}p{(\columnwidth - 8\tabcolsep) * \real{0.2000}}
  >{\raggedright\arraybackslash}p{(\columnwidth - 8\tabcolsep) * \real{0.2000}}
  >{\raggedright\arraybackslash}p{(\columnwidth - 8\tabcolsep) * \real{0.2000}}
  >{\raggedright\arraybackslash}p{(\columnwidth - 8\tabcolsep) * \real{0.2000}}@{}}
\toprule
\begin{minipage}[b]{\linewidth}\raggedright
model
\end{minipage} & \begin{minipage}[b]{\linewidth}\raggedright
verb
\end{minipage} & \begin{minipage}[b]{\linewidth}\raggedright
noun
\end{minipage} & \begin{minipage}[b]{\linewidth}\raggedright
adj
\end{minipage} & \begin{minipage}[b]{\linewidth}\raggedright
adv
\end{minipage} \\
\midrule
\endhead
tf & happen, \textbf{let}, want, gon, think & time, people, man,
problem, guy & wrong, good, great, right, little & right, kind, maybe,
actually, probably \\
tfidf & happen, \textbf{let}, gon, want, think & problem, wrong, relief,
pleasure & wrong, great, good, right, able & right, kind, actually,
maybe \\
pmi & \textbf{-i'll}, privilege, irk, do--, to\_do\_with, transpire &
barbecu, technicality-, stooling, musve, mcmahon'll & DON'T, youtryto,
uriel, cess, exploitable & montero, mackowes, it'about, vega, weenie \\
\bottomrule
\end{longtable}

Appendix 3A. Top 5 collocations and use cases for \emph{tell you
what}(freq 23115, group E).

\begin{longtable}[]{@{}
  >{\raggedright\arraybackslash}p{(\columnwidth - 4\tabcolsep) * \real{0.3333}}
  >{\raggedright\arraybackslash}p{(\columnwidth - 4\tabcolsep) * \real{0.3333}}
  >{\raggedright\arraybackslash}p{(\columnwidth - 4\tabcolsep) * \real{0.3333}}@{}}
\toprule
\begin{minipage}[b]{\linewidth}\raggedright
oxford
\end{minipage} & \begin{minipage}[b]{\linewidth}\raggedright
cambridge
\end{minipage} & \begin{minipage}[b]{\linewidth}\raggedright
merriam webster
\end{minipage} \\
\midrule
\endhead
\emph{\textbf{I'll} tell you what--- \textbf{let's} stay in instead.} &
\emph{\textbf{I'll} tell you what - we'll split the money between us.} &
\emph{(\textbf{I'll}) Tell you what---I'll \textbf{let} you borrow the
car if you fill it up with gas.} \\
\bottomrule
\end{longtable}

Appendix 3B. The representative use cases of \emph{tell you what}(freq
23115, group E)

\end{document}